%% file: main.tex
\documentclass{article}

\usepackage[nonatbib, final]{neurips_2024}
\usepackage[square,numbers]{natbib}

\usepackage[utf8]{inputenc} %
\usepackage[T1]{fontenc}    %
\usepackage{hyperref}       %
\usepackage{url}            %
\usepackage{booktabs}       %
\usepackage{amsfonts}       %
\usepackage{amssymb}
\usepackage{nicefrac}       %
\usepackage{microtype}      %
\usepackage{tabularx}
\usepackage[table]{xcolor}
\usepackage{adjustbox}
\usepackage{multirow}
\usepackage{enumitem}
\usepackage{pifont}

\usepackage{amsmath}
\usepackage{amsthm}
\usepackage{bm}
\usepackage{bbm}
\usepackage{xspace}
\usepackage{comment}
\usepackage{wrapfig}
\usepackage{algorithm}
\usepackage{algpseudocode}
\usepackage{graphicx}
\usepackage{subcaption}
\usepackage{listings}
\usepackage{makecell}

\definecolor{light}{rgb}{0.68, 0.90, 0.77}
\definecolor{orange}{rgb}{0.93, 0.74, 0.60}
\definecolor{lightorange}{rgb}{1, 0.87, 0.68}
\definecolor{lightgreen}{rgb}{0.76, 0.88, 0.76}
\definecolor{lightgray}{rgb}{0.95, 0.95, 0.97}
\definecolor{lightred}{rgb}{0.92, 0.29, 0.36}
\definecolor{multiflowcolor}{rgb}{0.92, 0.88, 1}
\definecolor{lightcyan}{rgb}{0.424, 0.651, 0.804}
\definecolor{zerocolor}{rgb}{0.91, 1, 0.999}

\definecolor{codeblue}{rgb}{0.25, 0.5, 0.5}
\definecolor{codekw}{rgb}{0.35, 0.35, 0.75}

\lstdefinestyle{Pytorch}{
    language         = Python,
    backgroundcolor  = \color{white},
    basicstyle = \fontsize{8.0pt}{9pt}\selectfont\ttfamily\bfseries,
    columns          = fullflexible,
    breaklines       = true,
    captionpos       = b,
    commentstyle     = \fontsize{4pt}{4pt}\color{codeblue},
    keywordstyle     = \fontsize{4pt}{4pt}\color{codekw},
    morekeywords     = {augment, softmax, confidence\_filter, torch, argmax},
}

\newcommand{\vtau}{\boldsymbol{\mathbf{t}_{\text{ctx}}}}

\newcommand{\vx}{\mathbf{x}}
\newcommand{\vt}{\mathbf{t}}
\newcommand{\vz}{\mathbf{z}}

\DeclareMathOperator*{\argmax}{arg\,max}

\newtheorem{theorem}{Theorem}[section]

\newtheorem{proposition}[theorem]{Proposition}

\newcommand{\ie}{\emph{i.e.}\xspace}
\newcommand{\wrt}{\emph{w.r.t.}\xspace}
\newcommand{\eg}{\emph{e.g.}\xspace}

\newcommand{\method}{\textsc{Zero}\xspace}

\title{Frustratingly Easy Test-Time Adaptation of Vision-Language Models}

\author{%
  Matteo Farina\textsuperscript{1,}\thanks{Correspondence to: \texttt{m.farina@unitn.it}. Code at \url{https://github.com/FarinaMatteo/zero}.}\quad
  Gianni Franchi\textsuperscript{2}\quad
  Giovanni Iacca\textsuperscript{1}\quad \vspace{.1em}\\
  \textbf{Massimiliano Mancini\textsuperscript{1}}\quad
  \textbf{Elisa Ricci\textsuperscript{1,3}} \vspace{1em} \\
  \textsuperscript{1}University of Trento \vspace{.1em}\\ \textsuperscript{2}U2IS, ENSTA Paris, Institut Polytechnique de Paris \vspace{.1em}\\ \textsuperscript{3}Fondazione Bruno Kessler (FBK)
}

\begin{document}

\maketitle

\begin{abstract}
 Vision-Language Models seamlessly discriminate among arbitrary semantic categories, yet they still suffer from poor generalization when presented with challenging examples. For this reason, Episodic Test-Time Adaptation (TTA) strategies have recently emerged as powerful techniques to adapt VLMs in the presence of a single unlabeled image.
 The recent literature on TTA is dominated by
 the paradigm of prompt tuning by Marginal Entropy Minimization, which, relying on online backpropagation, inevitably slows down inference while increasing memory.
 In this work, we theoretically investigate the properties of this approach and unveil that a surprisingly strong TTA method lies dormant and hidden within it. 
 We term this approach \method (\textit{TTA with ``zero'' temperature}), whose design is both incredibly effective and frustratingly simple: augment $N$ times, predict, retain the most confident predictions, and marginalize after setting the Softmax temperature to zero.
 Remarkably, \method requires a \emph{single} batched forward pass through the vision encoder only and \emph{no} backward passes. 
 We thoroughly evaluate our approach following the experimental protocol established in the literature and show that \method largely surpasses or compares favorably \wrt the state-of-the-art while being almost $10\times$ faster and $13\times$ more memory friendly than standard Test-Time Prompt Tuning. 
 Thanks to its simplicity and comparatively negligible computation, \method can serve as a strong baseline for future work in this field.
 The code is available.
\end{abstract}

\section{Introduction}\label{sec:intro}
Groundbreaking achievements in Vision-Language pretraining~\cite{radford2021learning,jia2021scaling,samadh2023align,singh2022flava,zhai2023sigmoid} have increased the interest in crafting Vision-Language Models (VLMs) that can understand visual content alongside natural language, enabling a new definition of zero-shot classification. 
Despite huge pretraining databases~\cite{schuhmann2022laion, sharma2018conceptual}, VLMs still face limitations, suffering from performance degradation in case of large train-test dissimilarity~\cite{mayilvahanan2023does} and requiring the design of highly 
generalizing textual templates~\cite{zhou2022learning}.

Test-Time Adaptation (TTA) can effectively improve the robustness of VLMs by adapting a given model to online inputs. 
Among the various TTA setups (such as ``fully''~\cite{wang2020tent}, ``continual''~\cite{wang2022continual} or ``practical'' TTA~\cite{yuan2023robust}), Episodic TTA~\cite{zhang2022memo} is particularly appealing, as it focuses on \emph{one-sample} learning problems and requires no assumptions on the distribution of the test data.
When presented with a test image $\vx$, the parameters $\theta$ of a model $f$ are optimized through a TTA objective $\mathcal{L}$ before inferring the final prediction, and reset afterward.

The choice of $\mathcal{L}$ is, ultimately, what characterizes TTA methods the most, with the recent literature being dominated by the objective of Marginal Entropy Minimization (MEM) \cite{zhang2022memo}.
Given a collection $\mathcal{A}$ of $N \in \mathbb{N}$ data augmentation functions, a test image $\vx$ is first augmented $N$ times to obtain a set of different views $X = \{\mathcal{A}_i (\vx)\}_{i=1}^N=\{ \vx_i\}_{i=1}^N$.
The \emph{marginal probability distribution} $\overline{p}$ \wrt sample
$\vx$ is then defined as the empirical expectation of Softmax-normalized model outputs over $X$, \ie:
\begin{equation}
\overline{p}(\cdot|\vx) = \frac{1}{N} \sum_{i=1}^{N} p(\cdot|\vx_i).
\end{equation}
Under this framework, the Shannon Entropy of $\overline{p}$ is a bona fide measure of how \emph{inconsistently} and \emph{uncertainly} the model predicts over $X$, making it a tantalizing candidate to minimize, \ie:
\begin{equation}\label{equation:entMin}
 \mathcal{L}_{ent} = H(\overline{p}(\cdot|\vx)) =  -\sum_{c=1}^C \overline{p}(y=c|\vx) \log(\overline{p}(y=c | \vx)),
\end{equation}
where $C$ is the number of semantic categories. 
Once $\mathcal{L}_{ent}$ is computed, (some of) the parameters of $f$ are typically updated for a few steps of Gradient Descent before inferring the final prediction over the source input $\vx$ with updated parameters. 
Owing to its simplicity and effectiveness, MEM has become a \emph{de facto} standard in modern TTA~\cite{zhang2022memo, shu2022test, samadh2023align, lee2024entropy, sui2024just, niu2023towards}.

In this work, we take the opposite direction and challenge this paradigm. 
By conducting an in-depth theoretical and empirical investigation, we find that: \ding{172} while effective in improving model robustness, MEM has \emph{little effect} on the prediction of $\overline{p}$; \ding{173} no matter the dataset, the label space, or the parameter initialization, VLMs become much better classifiers when $\overline{p}$ replaces the standard inference protocol.
Building on these insights, we show that a surprisingly strong and optimization-free TTA baseline is subtly hidden within the MEM framework.  
We term this baseline \method, which is short for TTA with ``\textbf{zero}'' temperature.
Instead of tuning any parameters, setting to zero the Softmax temperature before marginalizing over views makes $\overline{p}$ already stronger than the model after MEM. 
Notably, \method only requires a single forward pass through the vision encoder and no backward passes.

Wrapping up, the contributions of this paper are the following:
\begin{enumerate}[leftmargin=2em]
    \item We theoretically show \emph{when} the prediction obtained through $\bar{p}$ (\ie, $\argmax \overline{p}$) is \emph{invariant} to MEM, and empirically verify that MEM has largely \emph{no effect} on $\argmax \overline{p}$; 
    
    \item We theoretically and empirically demonstrate that the error rate of $\overline{p}$ is a lower bound to the base error of a VLM in the setup of TTA. 
    Additionally, we identify augmentations-induced \emph{overconfidence} as the primal factor undermining the reliability of $\overline{p}$;

    \item Motivated by these theoretical insights, we introduce \method, a frustratingly simple TTA approach that recovers the reliability of $\overline{p}$ by tweaking a single parameter of the model: the temperature; 

    \item We thoroughly evaluate \method following the established experimental setup with a variety of model initializations. Our results show that \method surpasses or compares favorably to state-of-the-art TTA methods while being much faster and more memory efficient (\eg, $10\times$ faster and $13\times$ more memory efficient than the established Test-Time Prompt Tuning \cite{shu2022test}).
\end{enumerate}

\section{Understanding Marginal Entropy Minimization}
\label{sec:understanding}
In this Section, we take a step towards both theoretically and empirically understanding the paradigm of MEM.   
In particular, this section is devoted to answering the following research questions: 
\begin{enumerate}
[leftmargin=2em]
    \item \emph{How does MEM affect the marginal probability distribution?} And, in turn
    \item \emph{How does the marginal probability distribution relate to the standard inference protocol?}
\end{enumerate}
First, we introduce MEM for VLMs by reviewing the established Test-Time Prompt Tuning (TPT) method~\cite{shu2022test} and its notation.
Then, in Sections~\ref{sec:entInvariance} and \ref{sec:margBound} we answer the research questions above. 

\subsection{Preliminaries}
\textbf{Zero-Shot Classification with VLMs} employs a predefined template (\eg, ``\texttt{a photo of a}'') from which a set of context vectors $\vtau$ is obtained by looking up a token embedding table.
Expanding the template with the class names (\eg, ``\texttt{a photo of a laptop.}'' for the class ``\texttt{laptop}'') makes up the entire set of input vectors $[\vtau, \vt_1], \ldots, [\vtau, \vt_C]$, with $\vt_i$ being the embeddings derived from the $i$-th class name.
A text encoder $\mathbf{E}_{\text{txt}}$ transforms these class descriptions into independent and normalized text embeddings $\vz_1^{\text{txt}}, \ldots, \vz_C^{\text{txt}}$ and an image encoder $\mathbf{E}_{\text{img}}$ encodes an input image $\vx$ into a normalized latent vector $\vz^{\text{img}}$.
Lastly, classification is carried out by picking the class $c$ corresponding to the text embedding $\vz_{c}^{\text{txt}}$ holding the maximum cosine similarity with $\vz^{\text{img}}$.

\textbf{MEM for VLMs.}
Pioneered by MEMO \cite{zhang2022memo} in the scope of unimodal neural networks, MEM was re-purposed for TTA with VLMs by Test-Time Prompt Tuning \cite{shu2022test}.
In \cite{shu2022test}, a VLM such as CLIP~\cite{radford2021learning} is adapted at test time by minimizing the same objective of Eq.~\eqref{equation:entMin}.
In contrast to optimizing all model parameters, TPT relies on 
the effectiveness of prompt tuning \cite{zhou2022learning, zhou2022conditional, khattak2023maple}, optimizing only the context vectors derived from the token embeddings of the standard CLIP template 
``\texttt{a photo of a}''. 
By explicitly enunciating the dependency on the context vectors $\vtau$ and re-using the notation of Sec.~\ref{sec:intro}, one can re-write the MEM objective of \cite{shu2022test} as:  
\begin{equation}\label{equation:entMinTPT}
\begin{split}
 & \mathcal{L}_{ent} = H(\overline{p}(\cdot|\vx, \vtau)) = -\sum_c^C \overline{p}(y=c|\vx,\vtau, \tau) \log\,\left(\overline{p}(y=c|\vx, \vtau, \tau )\right) \\
 & \text{where } \overline{p}(y=c|\vx,\vtau, \tau) = \frac{1}{N} \sum_i^N \frac{\exp \left( \vz_i^{\text{img}} \cdot \vz_c^{\text{txt}}(\vtau)/\tau\right)}{\sum_k^C \exp \left( \vz_i^{\text{img}} \cdot \vz_k^{\text{txt}}(\vtau)/\tau\right)}.
\end{split}
\end{equation}
Here, $\tau$ is the \emph{temperature} of the Softmax operator.
In the rest of this section, we omit the dependency on $\tau$ for simplicity, writing $p(\cdot|\vx,\vtau)$.
Similarly to \cite{zhang2022memo}, the objective of Eq.~\eqref{equation:entMinTPT} is minimized for a single step of Gradient Descent to update the set of context vectors. 
The updated context vectors, denoted as $\vtau^*$, are then used to prompt the VLM and obtain the final prediction for $\vx$.
For any class $c$ this is simply $\vz_{\vx}^{\text{img}} \cdot \vz_c^{\text{txt}}(\vtau^*)$,  which is easily transformed into $p(y=c|\vx, \vtau^*)$ via Softmax.

\subsection{How does MEM affect the marginal probability distribution?}\label{sec:entInvariance}
The recent literature on TTA shows that minimizing $\mathcal{L}_{ent}$ significantly enhances the robustness of model outputs. 
However, the impact of this process on the marginal probability distribution $\overline{p}$ remains unclear.
We start with a straightforward hypothesis: due to its nature, minimizing $\mathcal{L}_{ent}$ tends to increase the probability of the most probable class of $\overline{p}(\cdot|\vx,\vtau)$.
More formally, denoting with $\hat{c}$ the prediction of $\overline{p}$
(\ie, $\hat{c} = \argmax \overline{p}(\cdot|\vx,\vtau)$), we hypothesize that $\overline{p}(y=\hat{c}|\vx,\vtau^*) > \overline{p}(y=\hat{c}|\vx,\vtau)$.
If this hypothesis is realized, it comes as a natural consequence that minimizing $\mathcal{L}_{ent}$ is unlikely to alter the prevailing class of $\overline{p}$, thus resulting in a consistent prediction pre- and post-TTA where $\argmax \overline{p}(\cdot|\vx,\vtau)=\argmax \overline{p}(\cdot|\vx,\vtau^*)$.

Hence, the first contribution of this work is to show that the prediction of the marginal probability distribution $\overline{p}$ is \emph{invariant} to Entropy Minimization under loose constraints on confidence and gradients. 
To lighten the notation of the proposition, let us first define the following function $g$:
\begin{equation}
g(c,\vz^{\text{img}},\vz_{1}^{\text{txt}}, \ldots, \vz_{C}^{\text{txt}})=   \frac{\exp \left(\vz^{\text{img}} \cdot  \vz_{c}^{\text{txt}}/\tau \right)}{ \sum_k^C \exp \left(\vz^{\text{img}} \cdot  \vz_k^{\text{txt}}/\tau \right)}
\end{equation}
\ie, the probability assigned to class $c$ given a latent image representation $\vz^{\text{img}}$ and class-wise text embeddings $\vz_1^{\text{txt}}, \ldots, \vz_C^{\text{txt}}$.
Additionally, let $\delta g(c,\vz^{\text{img}})$ be the negative variation incurred to the function $g$ when the context vectors $\vtau$ are updated through Entropy Minimization:
\begin{equation}
\delta g(c, \vz^{\text{img}}) = g(c, \vz^{\text{img}}, \vz_1^{\text{txt}}(\vtau), \ldots, \vz_C^{\text{txt}}(\vtau)) - g(c, \vz^{\text{img}}, \vz_1^{\text{txt}}(\vtau^*), \ldots, \vz_C^{\text{txt}}(\vtau^*))
\end{equation}
where, for clarity, the dependency of the text embeddings $\vz_1^{\text{txt}}, \ldots, \vz_C^{\text{txt}}$ on the context vectors (either $\vtau$ or $\vtau^*$) is explicit.
Using this notation, we can formalize the following proposition:

\begin{proposition}
Let $\vz_1^{\text{img}}, \ldots, \vz_N^{\text{img}}$ be the latent image representations resulting from the $N$ views and $\hat{c} = \argmax \overline{p}(\cdot|\vx, \vtau)$ be the initial prediction of the marginal probability distribution. 
If the entropy of $\overline{p}$ is minimized and $\overline{p}(y=\hat{c}|\vx,\vtau) > \frac{1}{N} \sum_{i=1}^N \delta g (\hat{c}, \vz_i^{\text{img}})$
then the prevalent class of $\overline{p}$ is invariant to MEM, \ie, $\argmax \overline{p}(\cdot|\vx, \vtau ) = \argmax \overline{p}(\cdot|\vx, \vtau^*)$.
\end{proposition}

In Appendix \ref{app:mem},  we provide a detailed proof of this proposition, highlighting that $\delta (\hat{c}, \vz^{\text{img}})$ is directly linked to the gradient \wrt the context vectors $\vtau$.
This relationship emerges when writing any post-update text embedding $\vz_c^{\text{txt}}(\vtau^*)$ as a function of its pre-update counterpart $\vz_c^{\text{txt}}(\vtau)$. 
Specifically, we can write $\vz_c^{\text{txt}}(\vtau^*) = \mathbf{E}_{\text{txt}}([\vtau - \lambda \nabla_{\vtau}(\mathcal{L}_{ent}), \vt_c])$, which is equivalent to $\mathbf{E}_{\text{txt}}([\vtau, \vt_c]) - \lambda \nabla_{\vtau}\mathbf{E}_{\text{txt}}([\vtau^*, \vt_c])^t\nabla_{\vtau}(\mathcal{L}_{ent})$ after a first-order Taylor Expansion around $\vtau^*$.
Consequently, the proposition holds by a condition relating confidence (through $\overline{p}(y=c|\vx,\vtau)$) and gradients (through $\delta (\hat{c}, \vz^{\text{img}})$).
Alongside the proof, Appendix \ref{app:mem} presents evidence supporting this proposition for CLIP \cite{radford2021learning} on the ImageNet-1k validation set \cite{deng2009imagenet}, as well as across various datasets for natural distribution shifts: ImageNet-A \cite{hendrycks2021natural}, ImageNet-R \cite{hendrycks2021many}, ImageNet-v2 \cite{recht2019imagenet}, and ImageNet-Sketch \cite{wang2019learning}.

\subsection{How does $\overline{p}$ relate to the standard inference protocol?}\label{sec:margBound}

From prior work on Test-Time Augmentations (TTAug) with unimodal neural networks~\cite{son2023efficient, shanmugam2021better}, empirical evidence suggests that $\overline{p}(\cdot |\vx)$ is more robust than $p(\cdot |\vx)$. 
This observation leads to the hypothesis that the expected risk of predicting with $\overline{p}$ is lower than that of doing so with $p$. 
However, the literature lacks guarantees for this hypothesis, except for the peculiar case in which the risk function is the squared error, \ie, $\ell(a,b)=(a-b)^2$ \cite{kimura2021understanding}.\footnote{In Appendix \ref{app:risk}, we show that this bound generalizes to any function $\ell$ satisfying the triangular inequality.}

As the second contribution of this study, we show that the error rate of $\overline{p}(\cdot| \vx)$ does indeed lower-bound the error rate of $p(\cdot| \vx)$. 
We do so by revisiting the theory of model ensembling, and showing that analogous ideas can emerge for TTA.

\textbf{Preliminaries on model ensembling.} From the theory of classifier ensembling~\cite{kuncheva2014combining}, we know that if $f_1, \ldots, f_N$ are $N \in \mathbb{N}$ independent classifiers with error rate $\epsilon$ and $\vx$ is an example whose label is $y \in \{0,1\}$, then the probability that any group of $k$ classifiers picks the same \emph{wrong} label $f_i(\vx) = \hat{y} \neq y$ can be expressed with a Binomial distribution wrapping $N$ Bernoulli processes:
\begin{equation}\label{eq:majErr}
    P_{\hat{y}\neq y}(k) = \binom{N}{k}\epsilon^k(1-\epsilon)^{(N-k)}
\end{equation}

\textbf{Revisiting model ensembling for TTA.} Eq.~\eqref{eq:majErr} holds as long as all events modeled as Bernoulli processes are independent.
Thus, we have an equivalent error estimate for the setup in which only a single classifier $f$ is present and  $X_y = \{\vx_i\}_{i=1}^N$  is a set of independent examples with the same underlying label $y$. 
Within this framework, any group of $k$ examples in $X_y$ to which the classifier has assigned the same label $\hat{y}$ is also a set of independent Bernoulli processes, whose error probability is still quantified via Eq.~\eqref{eq:majErr}. 
Note that this resembles the TTA setup in the presence of $N$ views of the source sample $\vx$, as long as augmentations do not change their underlying labels. 
We refer the reader to Appendix \ref{app:independence} for a discussion about the independence assumption among different views.

\textbf{$\overline{p}$ is better than $p$ (if $f$ is calibrated).} The final step can be taken through the lens of \emph{model calibration} \cite{guo2017calibration}, a property requiring that the confidence of a classifier matches its accuracy.
For example, a calibrated classifier $f$ whose confidence is $0.7$ is expected to be correct $70\%$ of the times.
In the previous discussion, if we denote with $k(y)$ the number of examples correctly labeled as $y$, then the accuracy of the classifier is exactly $k(y)/N$.
It follows that there is a positive correlation between accuracy and confidence if $f$ exhibits good calibration, i.e., $\uparrow k(y)/N \implies \uparrow \overline{p}(y)$. %
Thus, the probability of picking the wrong class with this marginal probability is approximated by Eq.~\eqref{eq:majErr}. 
Given this relationship, we have that $\overline{p}(y) = \max \overline{p}(\cdot)$ if $k(y)$ matches or exceeds the majority within $N$. Thus, the probability of picking the wrong class with $\overline{p}$ is approximated by marginalizing out all values of $k$ that satisfy this criterion, which entails that the error or $\overline{p}$ can be expressed with the cumulative distribution of \eqref{eq:majErr}:
\begin{equation}\label{eq:cumMajErr}
    P_{\hat{y}\neq y}(\overline{p}) = \sum_{k=\lfloor N/2+1 \rfloor}^N \binom{N}{k}\epsilon^k(1-\epsilon)^{(N-k)}
\end{equation}
From the Condorcet Jury Theorem \cite{shapley1984optimizing}, we know that Eq.~\eqref{eq:cumMajErr} is a \emph{monotonically decreasing function} if the error $\epsilon$ is better than random guessing, which is likely to be the case for VLMs pretrained on a massive amount of web data such as CLIP.
Hence, we conclude that the error of $\overline{p}$ is a realistic lower bound for the base model error $\epsilon$ \emph{over a set of independent data points sharing the same label}. 

\textbf{Does this lower bound empirically realize?} 
We evaluate if the error of $\overline{p}$ consistently lower bounds the error of $p$ also in practical use cases, where model calibration is unknown and the label space is large. 
For this, we use \texttt{CLIP-ViT-B-16}~\cite{dosovitskiy2020image}, the ImageNet validation set, and four datasets reflecting Natural Distribution Shifts~\cite{hendrycks2021many,hendrycks2021natural,recht2019imagenet,wang2019learning}.
For all classes in each dataset, we first draw all images sharing the same label ($X_y$).  
Then, we compute the expected error $\epsilon(y)$ of the model on this subset, together with the error of $\overline{p}$ (ideally, Eq.~\eqref{eq:majErr}).
Lastly, we average these errors over the entire label space $\mathcal{Y}$.
We do \underline{not} restrict to the cases where $y$ is supported by the majority and we do \underline{not} re-organize predictions in a \emph{one-versus-all} scheme.
Fig.~\ref{fig:preliminary}(a) clearly shows that the error of $\overline{p}$ is a lower bound to the base error of the model also in practical use cases where the label space is large and guarantees on model calibration are possibly missing.
Importantly, this phenomenon persists \emph{no matter the dataset}.

\input{tables/preliminary}
\section{Simple and surprisingly strong TTA (for free)}
\label{sec:baseline}

The main point of Section~\ref{sec:entInvariance} is that MEM generally does not affect the predominant class of the marginal probability distribution $\overline{p}$. 
On the other hand, from Section~\ref{sec:margBound} one can conclude that through $\overline{p}$ the model becomes a much stronger classifier. Summarizing:
\begin{equation}
\begin{split}
    & \text{From Section~\ref{sec:entInvariance}: } \argmax\left( \overline{p}(\cdot|\vx, \vtau)\right) = \argmax\left( \overline{p}(\cdot|\vx, \vtau^*)\right) \\
    & \text{From Section~\ref{sec:margBound}: } 
    \begin{cases}
        P_{\hat{y}\neq y}(\overline{p}(\cdot |\vx, \vtau)) \leq P_{\hat{y}\neq y}(p(\cdot|\vx,\vtau)) \\
        \text{ and, equivalently} \\ 
        P_{\hat{y}\neq y}(\overline{p}(\cdot |\vx, \vtau^*)) \leq P_{\hat{y}\neq y}(p(\cdot|\vx,\vtau^*))\\
    \end{cases}
\end{split}
\end{equation}

Chaining observations together, it emerges that:
\begin{equation}
\begin{split}
    P_{\hat{y}\neq y}(p(\cdot|\vx, \vtau^*)) \geq P_{\hat{y}\neq y}(\overline{p}(\cdot|\vx, \vtau^*)) = P_{\hat{y}\neq y}(\overline{p}(\cdot|\vx, \vtau)) \\
\end{split}
\end{equation}
\ie, if all assumptions are met, the error of MEM $\geq$ error of $\overline{p}$ after MEM $=$ error of $\overline{p}$ \emph{without} {MEM}.
All in all, this TTA framework is hiding a surprisingly strong and optimization-free baseline: $\overline{p}$!
Next, we highlight the detrimental impact of data augmentations on this marginal probability distribution and introduce a simple trick to recover its reliability: \emph{zeroing-out} the Softmax temperature.

\subsection{Augmentations undermine the reliability of $\overline{p}$}\label{sec:overconfidence}
While augmentations are essential in TTA to obtain multiple views of the test instance, noisy views may constitute Out-of-Distribution (OOD) data, thus having the undesired effect of un-calibrating the model. 
To sidestep this issue, one can attempt to discriminate between in-distribution (\wrt to the pretraining data) and OOD views. 
Given that low confidence is a common trait in OOD data, a viable way to discriminate is confidence-based filtering, such as in TPT \cite{shu2022test}.
Formally, a smaller set of confident views are obtained following $X_{filt} = \{ \vx_i \in X | H(p(\cdot|\vx_i, \vtau)) < \rho \}$, where $\rho$ is a threshold retaining the views whose entropy is in the bottom-10\% percentile (lowest entropy). 
Despite its effectiveness, this filter cannot help when the reliability of $\overline{p}$ is undermined by \emph{overconfidence}. 

\textbf{Augmentations lead to poor calibration.}  We demonstrate the impact of augmentation-induced overconfidence using the same model and datasets of Section~\ref{sec:margBound}. 
For each dataset, we generate an augmented counterpart following the augmentation and filtering setup of TPT \cite{shu2022test}, \ie: we augment an input $N=64$ times using simple random resized crops and horizontal flips. Then, we only retain 10\% of the $N$ views according to confidence-based filtering, resulting in 6 views per sample.
Consequently, each augmented dataset contains $6\times$ more data points than its plain counterpart.
The Expected Calibration Error (ECE) \cite{guo2017calibration} reported in Appendix \ref{app:overconfidence} conveys that \ding{172} zero-shot CLIP is well-calibrated (ECE$~<0.1$ for all datasets), strongly supporting the theory of Section~\ref{sec:margBound} and \ding{173} \emph{the augmented visual space greatly increases the calibration error}.

\textbf{Poor calibration is frequently linked to overconfidence.} We investigate the reason for the increase in ECE by presenting reliability diagrams for the ImageNet validation set in Fig.~\ref{fig:preliminary}(b). 
In a reliability diagram, every bar below the identity line $y=x$ signals overconfidence (\ie, the confidence on the x-axis prevails over the accuracy on the y-axis), while the opposite signals under-confidence.
Notably, in the scope of our experiments, overconfidence is the primal factor leading to an increase in the ECE.
The error rate, in contrast, decreases slightly.
In Appendix \ref{app:overconfidence}, we also experiment across all datasets for Natural Distribution Shifts and different CLIP models pretrained on the 2B subset of LAION \cite{cherti2023reproducible, schuhmann2022laion}. 
Importantly, this phenomenon further persists within this extended experimental suite.

\subsection{\method: Test-Time Adaptation with ``zero'' temperature}
Since its reliability is severely undermined by augmentations-induced overconfidence, directly predicting through $\overline{p}$ is not an enticing baseline for TTA.
Concurrently, we also know that the error rate does not decrease when predicting over the augmented visual space.
Hence, we are interested in finding an efficient way to capitalize on these observations: relying on the predictions over the views, %
while %
ignoring potentially misleading confidence information. 
The key is to note that both desiderata are obtained by explicitly tweaking a single parameter of the model: \emph{the temperature}.
Specifically, setting the temperature to (the limit of) zero corresponds to converting probability distributions into one-hot encodings, hence exclusively relying on their $\argmax$ when marginalizing. 
Inspired by this idea we propose \method, Test-Time Adaptation with ``\textbf{zero}'' temperature.

\textbf{Procedure.} \method follows these simple steps: \ding{172} augment, \ding{173} predict, \ding{174} retain the most confident predictions, \ding{175} %
set the Softmax temperature to zero and \ding{176} marginalize.
The final prediction is the $\argmax$ of the marginal probability distribution computed after ``zeroing-out'' the temperature, \ie:
\begin{equation}
    \label{eq:zero}
    \method(\vx, \vtau, C) = \argmax_{c \in [1,\ldots,C]} \left( \sum_{i=1}^N \mathbbm{1}(\vx_i \in X_{filt}) \lim_{\tau \to 0^{+}}{p}(y=c|\vx_i,\vtau, \tau) \right),
\end{equation}
where $\mathbbm{1}$ is an indicator function, whose output is $1$ if $\vx_i \in X_{filt}$ and $0$ otherwise, and $X_{filt}$ is the set of confident views \emph{before} tweaking the temperature, \ie, $\vx_i\in X_{filt}$ if $H(p(\cdot|\vx_i, \vtau, \tau) < \rho$.

\input{tables/torch}\textbf{Efficient Implementation.} In all its simplicity, \method is computationally lightweight.
In closed set assumptions where the class descriptions (and thus their embeddings) are fixed, \method only requires a single batched forward pass through the vision encoder, just as much as needed to forward the $N$ views.
Additionally, since the temperature is explicitly tweaked, \method needs \emph{no backpropagation at all} and can be implemented in a few lines of code. 
For reference, a PyTorch-like implementation \cite{paszke2019pytorch} is reported in Algorithm \ref{alg:zero-pytorch}.

\textbf{Equivalent perspective and final remark.} We bring to attention a simple scheme which corresponds to \method: \emph{voting} over (confident) augmentations. 
Drawing from the theory of ensembling, %
note that the error rate of the voting paradigm is exactly described by Eq.~\eqref{eq:majErr}.
Essentially, this means that \method capitalizes on the theoretical insights while circumventing practical issues stemming from augmentations.
We also highlight that \method is subtly hidden within any TTA framework relying exclusively on MEM, since computing $\overline{p}$ is inevitable therein.
For this reason, we refer to \method as a \emph{baseline} for TTA.
Our goal diverges from introducing a ``novel'' state-of-the-art method for TTA.
In contrast, we advocate the importance of evaluating simple baselines.

\section{Experiments}\label{sec:exps}
\input{tables/cr/nds}

In this section, we present a comprehensive experimental evaluation of \method.
Similarly to \cite{shu2022test, samadh2023align, zhao2024testtime}, we always work in the setup of \emph{single test point} adaptation.
Our results show that \method, alongside its simplicity, is an effective and efficient approach for TTA.

\subsection{Experimental Protocol}

\textbf{Baselines.} 
We compare \method to three strategies for TTA with VLMs: \ding{172} TPT \cite{shu2022test}, \ding{173} PromptAlign \cite{samadh2023align}, and \ding{174} Reinforcement Learning from CLIP Feedback (RLCF) \cite{zhao2024testtime}.
As introduced in Section~\ref{sec:understanding}, TPT works by minimizing the entropy of $\overline{p}$. 
In contrast, PromptAlign relies on a pretrained \texttt{MaPLe} initialization \cite{khattak2023maple} and pairs the MEM objective with a distribution alignment loss between layer-wise statistics encountered online and pretraining statistics computed offline.
Finally, RLCF does not include MEM in its framework; \citet{zhao2024testtime} shows that, if rewarded with feedback from a stronger teacher such as \texttt{CLIP-ViT-L-14}, the smaller \texttt{CLIP-ViT-B-16} can surpass the teacher itself.

\textbf{Models.} As different approaches consider different backbones in the original papers, we construct different comparison groups to ensure fair comparisons with all TTA baselines \cite{shu2022test,samadh2023align,zhao2024testtime}.

\emph{Group 1:} When comparing to TPT, we always use \texttt{CLIP-ViT-B-16}.
\citet{shu2022test} also reports \texttt{CLIP-Ensemble}, \ie, \texttt{CLIP} enriched with an ensemble of hand-crafted prompts.
While the design of TPT does not allow leveraging text ensembles (as also pointed out by concurrent work \cite{sui2024just}), \method seamlessly integrates with \texttt{CLIP-Ensemble}.
We denote this variant with \textsc{Zero}\textsubscript{+Ensemble}.

\emph{Group 2:} When comparing to PromptAlign, we follow \citet{samadh2023align} and start from a \texttt{MaPLe} initialization for a fair comparison.
\texttt{MaPLe} prompts are learned on ImageNet, following \cite{samadh2023align}.
Within this group, we also report TPT on top of \texttt{MaPLe}, as in \cite{samadh2023align}.

\emph{Group 3:} When comparing to RLCF, we use both \texttt{CLIP-ViT-B-16} and \texttt{CLIP-ViT-L-14} as in \cite{zhao2024testtime}.
Specifically, confidence-based filtering acts on top of the output of the first model, and the selected inputs are passed to the second model for the final output.
Both forward passes are inevitable in RLCF, so this scheme corresponds to ``early-exiting'' the pipeline, exactly as per MEM.
RLCF can vary according to (i) the parameter group being optimized and (ii) the number of adaptation steps.
We denote with $\Theta_v$ the full image encoder tuning, with $\vtau$ prompt tuning, and with $t$ the number of adaptation steps. 
For example, RLCF~$_{t=3}^{\vtau}$ indicates RLCF with prompt tuning for 3 TTA steps.
Note that, since all methods need to forward more than one image to the teacher model, the zero-shot baseline of this group is exactly zero-shot classification with \texttt{CLIP-ViT-L-14}.

\textbf{Pretrainings.} This Section deals with models officially released by OpenAI \cite{openai2024clip}. 
Appendix \ref{app:more_exp} further reports experiments with LAION-pretrained CLIP models \cite{cherti2023reproducible}, as well as the soft prompt initialization with supervised Context Optimization (\texttt{CoOp}) from \cite{zhou2022learning}.

\textbf{Benchmarks.} We follow the established experimental setup of \cite{shu2022test, samadh2023align}, evaluating \method on Natural Distribution Shifts and Fine-grained Classification (also referred to as ``Cross-Datasets Generalization" in previous works).
For the former, we consider the ImageNet validation set and the four datasets for Natural Distribution Shifts already presented in Section~\ref{sec:understanding}, commonly considered Out-of-Distribution (OOD) datasets for CLIP. For fine-grained classification, we evaluate all TTA methods on 10 datasets. %
Specifically, we experiment with Oxford-Flowers (FWLR) \cite{nilsback2008automated}, Describable Textures (DTD) \cite{cimpoi2014describing}, Oxford-Pets (PETS) \cite{parkhi2012cats}, Stanford Cars (CARS) \cite{krause20133d}, UCF101 (UCF) \cite{soomro2012ucf101}, Caltech101 (CAL)\cite{fei2004learning}, Food101 (FOOD) \cite{bossard2014food}, SUN397 (SUN)\cite{xiao2010sun}, FGVC-Aircraft (AIR) \cite{maji2013fine} and EuroSAT (ESAT) \cite{helber2019eurosat}.
For all of these datasets, we refer to the test split in \citet{zhou2022learning} as per the common protocol. 

\textbf{Textual prompts.} 
When \texttt{+Ensemble} is specified, we do \emph{not} use dataset-specific templates. 
In contrast, we use the set of 7 generic templates highlighted in the official CLIP repository \cite{openai2024clip} across all datasets. 
When adapting \texttt{MaPLe}, we stick to the ImageNet-learned prompts released by \cite{khattak2023maple} and evaluate them cross-datasets as in \cite{samadh2023align}.

\textbf{Implementation Details.} The augmentation pool $\mathcal{A}$ only contains random resized crops and random horizontal flips. 
The only hyperparameter of \method is the percentile for confidence-based filtering, which is set to $0.3$ after validation on ImageNet (following standard practice \cite{zanella2024test}) and kept fixed \emph{for all datasets}.  
We inherit the setup of TPT with $N=64$, crafting $63$ augmentations to collate with the source image.
To ensure hardware differences do not play any role in comparisons, we execute all TTA methods under the same hardware setup by running the source code of each repository with no modifications.
We always use 1 NVIDIA A100 GPU and FP16 Automatic Mixed Precision.
Results are averaged over 3 different seeds.
Unless otherwise specified, all tables report top-1 accuracy.

\subsection{Results}

\textbf{Natural Distribution Shifts.} 
Results for Natural Distribution Shifts are reported in Table \ref{tab:nds}.

\emph{Group 1 (TPT):} \method \emph{surpasses TPT consistently on all datasets}.
Among OOD datasets, peak difference is reached with ImageNet-A, where \method outperforms TPT by $+4.84\%$.
Enriching \method with hand-crafted prompts improves results further, with an average margin of $+3.66\%$ \wrt TPT.

\emph{Group 2 (PromptAlign):} Within the second comparison group, \method \emph{outperforms PromptAlign on all datasets}, with $+1.68\%$ being the gap in average performance.
 \method consistently outperforms TPT also when the baseline initialization is \texttt{MaPLe} (by an average of $+2.92\%$).
Please note that we omit evaluation on ImageNet for this group, since PromptAlign adopts token-level statistics from this dataset when adapting to test points, which would render the comparison unfair.
For completeness, we report that zero-shot \texttt{MaPLe} achieves an accuracy of $70.72\%$ on ImageNet, which is improved to $72.99\%$ by adapting with \method ($+2.27\%$).

\emph{Group 3 (RLCF):} 
We follow \cite{zhao2024testtime} and report RLCF variants with $t=3$ steps. 
In this group, \method outperforms RLCF in 5 out of 5 datasets, with a gap in the average performance of $+1.25\%$.
Importantly, RLCF is only close to \method with image encoder tuning; only prompt tuning is insufficient.

\textbf{Fine-grained Classification.}
\input{tables/cr/fg}
Results for fine-grained classification are shown in Table \ref{tab:fc}.
To foster readability, the standard deviations of \method are separately reported in Table \ref{tab:std} (Appendix).

\emph{Group 1 (TPT):} Default \method improves over the zero-shot baseline \texttt{CLIP-ViT-B-16}, but is outperformed by TPT with an average margin of $-0.57\%$. 
However, extending \method with hand-crafted prompts (something that TPT cannot do \emph{by design}) is sufficient to outperform TPT on 7 out of 10 datasets, and obtain an average improvement of $+0.74\%$. 

\emph{Group 2 (PromptAlign):} On average, PromptAlign has an improvement of $+0.5\%$ over \method.
However, note that this is mostly influenced by the performance on one dataset only (EuroSAT) and that, in contrast, \method \emph{surpasses PromptAlign in 7 out of 10 datasets}.
In line with the previous benchmark, \method better adapts \texttt{MaPLe} than TPT, again outperforming it in 7 out of 10 datasets.

\emph{Group 3 (RLCF):} As \citet{zhao2024testtime} do not report results on fine-grained classification, we use their code %
to evaluate four RLCF variants: $\Theta_v$ and $\vtau$ tuning, with $t=1$ and $t=3$ adaptation steps.
We find that \method largely outperforms RLCF regardless of the configuration.
Even with respect to the strongest RLCF~$_{t=3}^{\Theta_v}$ variant, \method obtains an average improvement of $+2.28\%$.

\textbf{Computational Requirements.}
The complexity of \method does not scale linearly with the size of the label space, as it does for prompt-tuning strategies.
To quantify the computational gain of \method \wrt other TTA methods, we report the runtime per image and peak GPU memory in Table \ref{tab:perf} under the same hardware (\ie, 1 NVIDIA RTX 4090). 
We compare the computational requirements of \method to TPT and the RLCF pipeline in a worst-case scenario where the label space is large (ImageNet). 
We omit PromptAlign from our analysis since it has slightly worse computational performance than TPT.

\method is $9.5\times$ faster than TPT taking $12.61\times$ less memory, corresponding to an order of magnitude of computational savings in both time and space. \input{tables/perf}
Concerning the slowest RLCF variant (prompt tuning), \method is $15\times$ faster and takes $7.22\times$ less memory.
In the faster RLCF~$^{\Theta_v}$, text classifiers are also cached; nevertheless, \method is $2.25\times$ faster and $3.5\times$ more memory friendly.

\section{Related Work}
Closest to our work is a recent and very active research thread focusing on Episodic TTA with VLMs \cite{shu2022test, samadh2023align, zhao2024testtime, sui2024just}.
As discussed in the manuscript, these methods mostly rely on prompt learning, a parameter-efficient strategy that only trains over a small set of input context vectors \cite{li2021prefix}.
Narrowing down to VLMs, notable examples of prompt learning approaches include CoOp \cite{zhou2022learning}, CoCoOp \cite{zhou2022conditional}, and MaPLe \cite{khattak2023maple}.
Episodic TTA has also been explored with traditional unimodal networks, such as ResNets \cite{he2016deep}, where MEM is still a core component \cite{zhang2022memo}.
In this context, MEM has recently been enriched with sharpness- \cite{niu2023towards} or shape-aware filtering \cite{lee2024entropy}. 
Due to its nature, Episodic TTA is completely agnostic to the temporal dimension and is powerful when no reliable assumptions on the test data can be taken.
Some other works relax these constraints and integrate additional assumptions such as \emph{batches} of test data being available instead of single test points \cite{wang2020tent}.
When test data are assumed to belong to the same domain, one can rely on various forms of knowledge retention as a powerful mechanism to gradually incorporate domain knowledge \cite{liu2024dart, ma2024swapprompt} or avoid forgetting \cite{niu2022efficient}.
The synergy between TTA and retrieval is also emerging as a powerful paradigm when provided with access to huge external databases \cite{hardt2024testtime, zancato2023train}. 
We particularly believe this can be a promising direction.

Closely related to our work are also Test-Time Training (TTT) and TTAug.
In TTT the same \emph{one sample} learning problem of Episodic TTA is tackled with auxiliary visual self-supervised tasks, such as rotation prediction \cite{sun2020test} or masked image modeling \cite{gandelsman2022test}, which require specialized architecture heads and are not directly applicable to VLMs.
TTAug has recently been theoretically studied \cite{kimura2021understanding}. It boils down to producing a large pool of augmentations to exploit at test time \cite{shanmugam2021better}, or to learn from \cite{tomar2023tesla}.
In all its simplicity, \method can be seen as a strong TTAug baseline for VLMs, which, differently from concurrent work \cite{zanella2024test}, does not involve any form of optimization.

\input{sections/limitations}

\section{Conclusions}
We theoretically investigated Marginal Entropy Minimization, the core paradigm of the current research on Test-Time Adaptation with VLMs.
Building on our theoretical insights, we introduced \method: a frustratingly simple yet strong baseline for TTA, which only relies on a single batched forward pass of the vision encoder.
\method works by setting the temperature of the Softmax operator to ``zero'' before marginalizing across confident views, which is equivalent, in terms of output, to the widely known paradigm of majority voting.
Our experimental results on Natural Distribution Shifts and Fine-grained Classification unveil that \method favorably compares to state-of-the-art TTA methods while requiring relatively negligible computation. 
We hope our findings will inspire researchers to push the boundaries of TTA further.
\clearpage

\paragraph{Acknowledgements.}
The authors acknowledge the CINECA award under the ISCRA initiative for the availability of high-performance computing resources and support. 
Matteo Farina is supported by the PRIN project ``LEGO-AI'' (Prot.2020TA3K9N) and the PAT project ``AI@TN".
This work was supported by the projects EU Horizon ELIAS (No. 101120237), AI4TRUST (No.101070190), FAIR - Future AI Research (PE00000013), funded by NextGeneration EU, and carried out in the Vision and Learning joint laboratory of Fondazione Bruno Kessler and the University of Trento, Italy.

\bibliographystyle{plainnat}  
\bibliography{refs} 

\clearpage
\include{sections/appendix}

\end{document}

%% file: tables/preliminary.tex
\begin{figure}[t]
    \centering
    \begin{tabular}{cc}
        \includegraphics[width=0.34\textwidth]{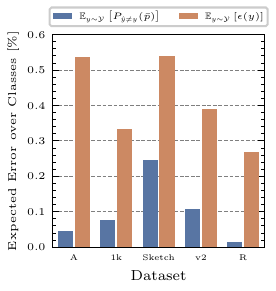} & \includegraphics[width=0.61\textwidth]{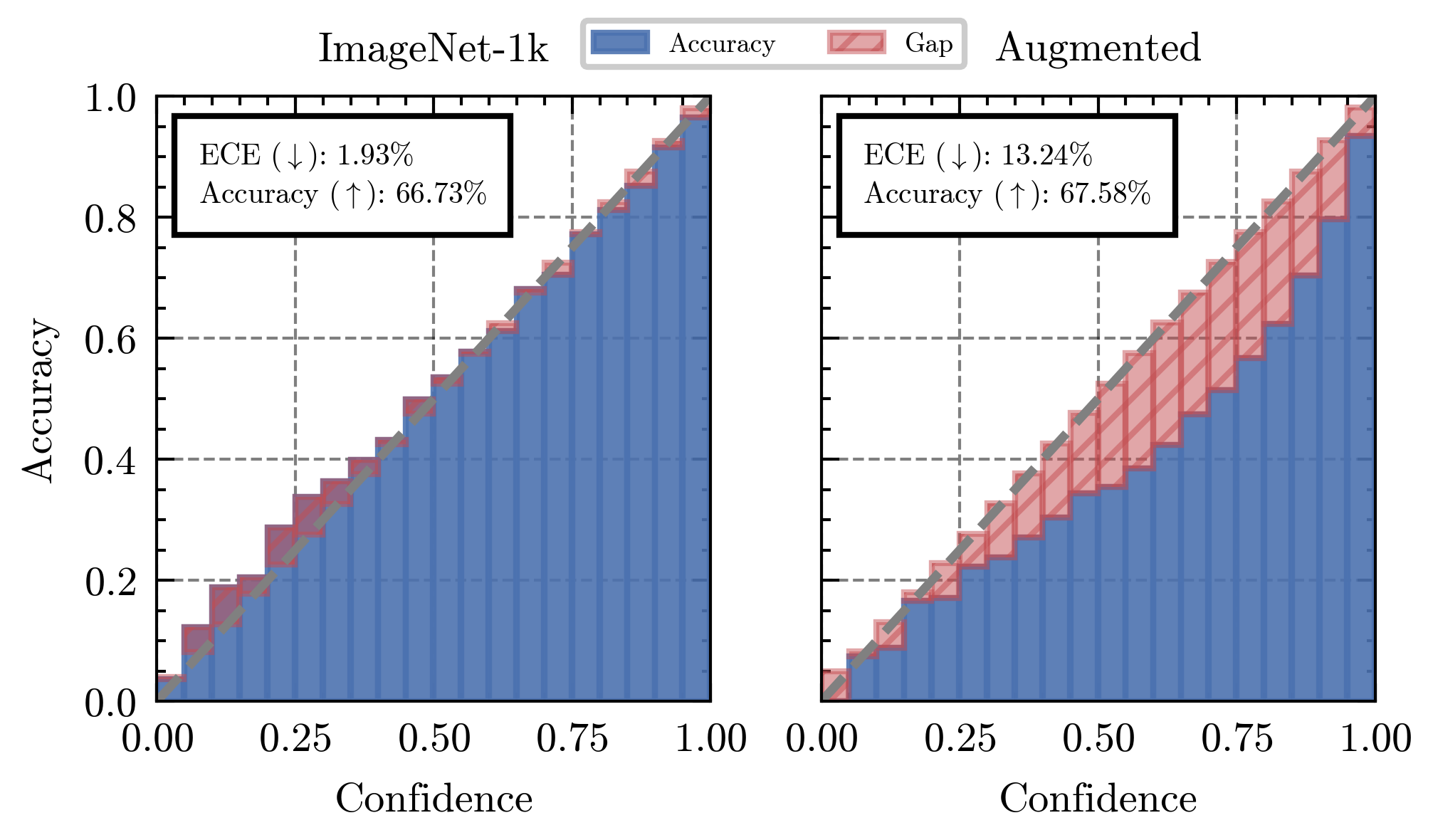} \\
        \small(a) Error of $\overline{p}$ vs $\epsilon(y)$. & \small(b) Reliability diagrams for IN-1k and its augmented version.
    \end{tabular}
    \caption{Motivating findings. (a) Comparison between the expected error of \texttt{CLIP-ViT-B-16}, denoted as $\epsilon(y)$, and the error of the marginal probability distribution obtained by marginalizing over examples with the same label, $P_{\hat{y}\neq y}(\overline{p})$; (b) Reliability diagrams of \texttt{CLIP-ViT-B-16} on the ImageNet validation set (left), and its augmented version (right), showing that augmentations largely un-calibrate CLIP exclusively due to overconfidence while leading to slightly better overall accuracy.}
    \label{fig:preliminary}
\end{figure}

%% file: tables/torch.tex
\begin{wrapfigure}{t}{0.52\textwidth}
\vspace{-5ex}
\begin{minipage}[t]{0.52\textwidth}
\begin{algorithm}[H]
\caption{PyTorch-style code for \method}
\label{alg:zero-pytorch}
\vspace{-1.ex}
\begin{lstlisting}[style=Pytorch,escapeinside={(@}{@)}]
# z_txt = pre-computed text embeddings (C,hdim)
# temp = model's original temp
# augment = takes (C,H,W) and returns (N,C,H,W)
# gamma = filtering percentile (e.g., 0.1)
def zero(image, z_txt, N, gamma, temp):
  # step1: augment
  views = augment(image, num_views=N)
  # step2: predict (unscaled logits)
  l = model.image_encoder(views) @ z_txt.t() 
  # step3: retain most confident preds
  l_filt = confidence_filter(l, temp, top=gamma) 
  # step4: zero temperature
  zero_temp = torch.finfo(l_filt.dtype).eps
  # step5: marginalize
  p_bar = (l_filt / zero_temp).softmax(1).sum(0)
  return p_bar.argmax()
\end{lstlisting}
\vspace{-1.ex}
\end{algorithm}
\end{minipage}
\vspace{-1.5em}
\end{wrapfigure}

%% file: tables/cr/nds.tex
\begin{table}[t!]
    \def\arraystretch{1.075}
    \caption{Natural Distribution Shifts. TTA methods are grouped according to the baseline model and top-1 accuracy is reported. \textbf{Bold text} is the best method within each group.}
    \centering
    \scriptsize
    \begin{tabularx}{\textwidth}{
     >{\raggedright\arraybackslash}p{1.8cm}
     >{\centering\arraybackslash}X
     >{\centering\arraybackslash}X
     >{\centering\arraybackslash}X
     >{\centering\arraybackslash}X
     >{\centering\arraybackslash}X
     >{\centering\arraybackslash}X
    }
     \toprule
    \textbf{Method} & ImageNet & A & V2 & R & Sketch & Mean \\

    \midrule
    \rowcolor{lightgray}
    \multicolumn{7}{c}{\texttt{CLIP-ViT-B-16}} \\
    [-2ex] \\ 

    \emph{Zero-Shot} & 66.73 & 47.87 & 60.86 & 73.98 & 46.09 & 59.11 \\
    \emph{Ensemble} & 68.34 & 49.89 & 61.88 & 77.65 & 48.24 & 61.20 \\
    TPT & 68.98 & 54.77 & 63.45 & 77.06 & 47.94 & 62.44 \\

    \rowcolor{zerocolor}
    \textsc{Zero} & 69.31$\pm$0.13& 59.61$\pm$0.19 & 64.16$\pm$0.03 & 77.22$\pm$0.05 & 48.40$\pm$0.07 & 63.74 \\

    \rowcolor{zerocolor}
    \textsc{Zero}\textsubscript{+Ensemble} & \textbf{71.17}$\pm$0.06 & \textbf{62.75}$\pm$0.14 & \textbf{65.23}$\pm$0.08 & \textbf{80.75}$\pm$0.02 & \textbf{50.59}$\pm$0.08 & \textbf{66.10} \\

    \cmidrule(lr){1-7}
    \rowcolor{lightgray}
    \multicolumn{7}{c}{\texttt{MaPLe}} \\
    [-2ex] \\
    \emph{Zero-Shot} & - & 50.90 & 64.07 & 76.98 & 49.15 & 60.28 \\
    TPT & - & 58.08 & 64.87 & 78.12 & 48.16 & 62.31 \\
    PromptAlign & - & 59.37 & 65.29 & 79.33 & 50.23 & 63.55 \\

    \rowcolor{zerocolor}
    \textsc{Zero} & - & \textbf{63.32}$\pm$0.26 & \textbf{66.81}$\pm$0.43 & \textbf{79.74}$\pm$0.32 & \textbf{51.07}$\pm$0.47 & \textbf{65.23} \\

    \cmidrule(lr){1-7}
    \rowcolor{lightgray}
    \multicolumn{7}{c}{\texttt{CLIP-ViT-B-16 + CLIP-ViT-L-14}} \\
    [-2ex] \\
    \emph{Zero-Shot} & 73.44 & 68.82 & 67.80 & 85.40 & 57.84 & 70.66 \\  
    RLCF~$_{t=3}^{\vtau}$ & 73.23 & 65.45 & 69.77 & 83.35 & 54.74 & 69.31 \\ [-2.5ex] \\
    RLCF~$_{t=3}^{\Theta_v}$ & 74.85 & 73.71 & 69.77 & 86.19 & 57.10 & 72.32 \\ [-2.5ex] \\

    \rowcolor{zerocolor}
    \textsc{Zero} & \textbf{75.52}$\pm$0.03 & \textbf{75.15}$\pm$0.26 & \textbf{70.37}$\pm$0.05 & \textbf{87.21}$\pm$0.09 & \textbf{59.61}$\pm$0.04 & \textbf{73.57} \\

     \bottomrule
    \end{tabularx}
    \label{tab:nds}
\end{table}

%% file: tables/cr/fg.tex
\begin{table}[t!]
    \def\arraystretch{1.15}
    \centering
    \caption{Fine-grained classification. TTA methods are grouped according to the reference baseline, top-1 accuracy is reported and \textbf{bold text} indicates the best performer of each group.}
    \scriptsize
    \begin{tabularx}{\textwidth}{
    >{\raggedright\arraybackslash}p{2cm}
    >{\centering\arraybackslash}X
    >{\centering\arraybackslash}X
    >{\centering\arraybackslash}X
    >{\centering\arraybackslash}X
    >{\centering\arraybackslash}X
    >{\centering\arraybackslash}X
    >{\centering\arraybackslash}X
    >{\centering\arraybackslash}X
    >{\centering\arraybackslash}X
    >{\centering\arraybackslash}X
    >{\centering\arraybackslash}X
    >{\centering\arraybackslash}X
    }
    
    \toprule
    \textbf{Method} & FLWR & DTD & PETS & CARS & UCF & CAL & FOOD & SUN & AIR & ESAT & Mean & Median \\

    \midrule
    \rowcolor{lightgray}
    \multicolumn{13}{c}{\texttt{CLIP-ViT-B-16}} \\ [-2.25ex] \\
     
    \emph{Zero-Shot} & 67.44 & 44.27 & 88.25 & 65.48 & 65.13 & 93.35 & 83.65 & 62.59 & 23.67 & 42.01 & 63.58 & 65.31 \\
     
    \emph{Ensemble} & 67.07 & 45.09 & \textbf{88.28} & 66.16 & 67.51 & 93.91 & 84.04 & 66.26 & 23.22 & \textbf{50.42} & 65.20 & 66.66 \\

    TPT & \textbf{68.75} & \textbf{47.04} & 87.23 & 66.68 & 68.16 & 93.93 & 84.67 & 65.39 & 23.13 & 42.86 & 64.78 & 67.42 \\

    \rowcolor{zerocolor}
    \textsc{Zero} & 67.68 & 46.12 & 87.75 & 68.04 & 67.77 & 93.66 & 86.53 &  65.03 & \textbf{25.21} & 34.33 & 64.21 & 67.72 \\

    \rowcolor{zerocolor}
    \textsc{Zero}\textsubscript{+Ensemble} & 67.17 & 45.86 & 87.83 & \textbf{68.97} & \textbf{69.18} & \textbf{94.41} & \textbf{86.77} & \textbf{67.63} & \textbf{25.21} & 42.17 & \textbf{65.52} & \textbf{68.30} \\

    \cmidrule(lr){1-13}
    \rowcolor{lightgray}
    \multicolumn{13}{c}{\texttt{MaPLe}} \\  [-2.25ex] \\
    
    \emph{Zero-Shot} & 72.23 & 46.49 & 90.49 & 65.57 & 68.69 & 93.53 & 86.20 & 67.01 & 24.74 & \textbf{48.06} & 66.30 & 67.85 \\  
    
    TPT & 72.37 & 45.87 & 90.72 & 66.50 & 69.19 & 93.59 & 86.64 & 67.54 & 24.70 & 47.80 & 66.49 & 68.36 \\
    
    PromptAlign & \textbf{72.39} & 47.24 & \textbf{90.76} & 68.50 & 69.47 & 94.01 & 86.65 & 67.54 & 24.80 & 47.86 & \textbf{66.92} & 68.99 \\

    \rowcolor{zerocolor}
    \textsc{Zero} & 71.62 & \textbf{47.89} & 90.60 & \textbf{68.58} & \textbf{69.87} & \textbf{94.48} & \textbf{87.20} & \textbf{68.20} & \textbf{26.25} & 39.47 & 66.42 & \textbf{69.23} \\

    \cmidrule(lr){1-13}
    \rowcolor{lightgray}
    \multicolumn{13}{c}{\texttt{CLIP-ViT-B-16 + CLIP-ViT-L-14}} \\ [-2.25ex] \\
     
    \emph{Zero-Shot} & 75.76 & 51.83 &  {92.86} &  {76.16} &  {73.70} & 94.04 & 88.03 & 66.96 & 30.54 & \textbf{54.38} & 70.43 & 74.73 \\
     
    RLCF~$_{t=1}^{\vtau}$ & 71.58 & 50.34 & 89.01 & 69.76 & 69.84 & 94.09 & 85.90 & 67.33 & 23.71 & 46.87 & 66.84 & 69.80 \\  
    
    RLCF~$_{t=3}^{\vtau}$ & 72.49 & 51.93 & 89.55 & 72.91 & 72.31 & 95.00 & 86.84 & 69.04 & 25.40 & 45.96 & 68.14 & 72.40 \\  
    
    RLCF~$_{t=1}^{\Theta_v}$ &  72.56 & 52.21 & 89.51 & 63.12 & 71.49 & 94.65 & 86.90 & 68.50 & 24.06 & 47.74 & 67.07 & 70.00 \\  
    
    RLCF~$_{t=3}^{\Theta_v}$ & 71.74 &  {53.27} & 91.15 & 70.93 & 73.24 &  {94.73} & 87.28 & 69.38 & 28.54 & 47.41 & 68.77 & 71.34 \\

    \rowcolor{zerocolor}
    \textsc{Zero} & \textbf{76.41} & \textbf{53.63} & \textbf{94.08} & \textbf{78.39} & \textbf{74.68} & \textbf{95.21} & \textbf{90.66} & \textbf{69.61} & \textbf{33.62} & 44.21 & \textbf{71.05} & \textbf{75.55} \\
    
    \bottomrule
    \end{tabularx}
    
    \label{tab:fc}
\end{table}

%% file: tables/perf.tex
\begin{table}[t]
\def\arraystretch{1.1}
    \caption{Computational requirements of different TTA methods.}
    \centering
    \scriptsize 
    \begin{tabularx}{0.9\textwidth}{
    >{\raggedright\arraybackslash}p{1.07cm}
     >{\centering\arraybackslash}X
     >{\centering\arraybackslash}X
     >{\centering\arraybackslash}X
     >{\centering\arraybackslash}X
     >{\centering\arraybackslash}X
    }
    \toprule

    \multirow{2.5}{*}{\textbf{Metric}} & \multicolumn{2}{c}{\texttt{CLIP-ViT-B-16}} & \multicolumn{3}{c}{\texttt{CLIP-ViT-B-16 + CLIP-ViT-L-14}} \\

    \cmidrule(lr){2-3}
    \cmidrule(lr){4-6}

    & TPT & \cellcolor{zerocolor}\textsc{Zero} & RLCF~$_{t=3}^{\vtau}$ & RLCF~$_{t=3}^{\Theta_v}$ & \cellcolor{zerocolor}\textsc{Zero} \\ 

    \cmidrule{1-6}

    Time [$s$] & 0.57$\pm0.01$ & \cellcolor{zerocolor}\textbf{0.06}$\pm0.01$ & 1.20$\pm0.02$ & 0.18$\pm0.01$ & \cellcolor{zerocolor}\textbf{0.08}$\pm0.02$ \\
    Mem [GB] & 17.66 & \cellcolor{zerocolor}\textbf{1.40} & 18.64 & 9.04 & \cellcolor{zerocolor}\textbf{2.58} \\

    \bottomrule

    \end{tabularx}
    \label{tab:perf}
\end{table}

%% file: sections/limitations.tex
\section{Limitations}
\method can seamlessly adapt a wide range of VLMs on arbitrary datasets without requiring extensive computational resources and is backed by theoretical justifications. 
However, we delineate four major limitations to our method which we report here. 

\textbf{Preliminary observations.} The first limitation concerns the preliminary observations which led to \method, such as augmentation-induced overconfidence or a comparable error rate between source and augmented datasets.
These observations may not persist if VLMs or benchmarks change significantly in the future, potentially leading to poor adaptation. 
For example, we have observed a consistent failure case for TTA with EuroSAT \cite{helber2019eurosat}, with \method incurring large performance drops \wrt simple zero-shot classification.
In Appendix \ref{app:eurosat} we unravel this worst-case further.%

\textbf{Theoretical assumptions.} The second limitation stems from theoretical assumptions, the core one being the invariance of the marginal probability distribution to marginal entropy minimization. 
While our proposition guarantees invariance if entropy is globally minimized and the negative variation to the probability of the most probable class is less than the initial probability itself, these theoretical assumptions may not hold all the time.
In this work, we supported our assumptions with empirical verification but, as per the first limitation, these may not extend to the space of all models and datasets. 
We refer the interested readers to Appendix \ref{app:mem} for a more in-depth discussion about the invariance of the prediction of $\overline{p}$ to MEM.

\textbf{Independence among views.}
A third worthy-of-note limitation relates to the independence assumption among the views from which the marginal probability distribution is obtained.
As we discussed in Section \ref{sec:margBound}, the views themselves do not have any \emph{direct} dependency, but they are still partially related through the source image from which they stem.
Related to this limitation, we hypothesize that extending \method in a Retrieval-Augmented TTA setup (or a cache-based one) could improve the results.
The discussion on this topic is extended in Appendix \ref{app:independence}.

\textbf{Linear complexity with respect to augmented views.} Finally, despite being much lighter than the current state-of-the-art TTA strategies, \method's computational requirements in the visual branch scale linearly with the number of views, since all of them need to be independently forwarded.
On this, we believe that exploring how to augment directly in the latent visual space to also circumvent the forward pass of the vision encoder is an intriguing direction.

%% file: sections/appendix.tex
\appendix 

\section{Marginal Entropy Minimization does not influence $\argmax \overline{p}$}\label{app:mem}

\subsection{Proof of Proposition 2.1.}
\begin{proof}

Let us denote the pre-TTA $p^{\text{init}}(c) = \overline{p}(y=c|\vx, \vtau)$ and the post-TTA $p^{\text{end}}(c)=\overline{p}(y=c|\vx, \vtau^*)$, \emph{i.e.}, the marginal probabilities before and after optimizing $\vtau$. 
Let $c^{\text{init}}$ and $c^{\text{end}}$ denote the predictions before and after TTA, \ie, $c^{\text{init}} = \argmax p^{\text{init}}$ and $c^{\text{end}} = \argmax p^{\text{end}}$.

To simplify the notation, let us use $\vz^{*\text{txt}}$ to write any post-TTA text embedding $\vz^{\text{txt}}(\vtau^*)$.

Under the assumption that entropy is minimized (the optimal scenario for MEM), we have $p^{\text{end}} (c^{\text{end}})=1$, and $p^{\text{end}} (c)=0 ~ \forall ~ c \neq c^{\text{end}}$.

Let us rewrite the final distribution $p^{\text{end}}$ using the function $g$ introduced in Sec.\ref{sec:entInvariance}. 
Specifically, for any class $c$, we have:
\begin{equation}
p^{\text{end}}(c)= \frac{1}{N} \sum_i^N \frac{\exp \left(\vz^{\text{img}}_i \cdot  \vz_{c}^{\text{txt}}(\vtau^*)/\tau \right)}{ \sum_k^C \exp \left(\vz^{\text{img}}_i \cdot  \vz_{k}^{\text{txt}}(\vtau^*)/\tau \right)} =  \frac{1}{N} \sum_i g(c,\vz^{\text{img}}_i ,\vz_{1}^{*\text{txt}}, \ldots, \vz_{C}^{*\text{txt}}) .
\end{equation}

Performing a first-order Taylor expansion on $g$, we have:
\begin{equation}\label{eq:gexpanded}
\begin{split}
&g(c,\vz^{\text{img}}_i ,\vz_{1}^{*\text{txt}}, \ldots, \vz_{C}^{*\text{txt}})
=g(c,\vz^{\text{img}}_i ,\vz_{1}^{\text{txt}}, \ldots, \vz_{C}^{\text{txt}}) + \\ 
&(\nabla_{[\vz_{1}^{\text{txt}}, \ldots, \vz_{C}^{\text{txt}} ]}  g )^t ([\vz_{1}^{*\text{txt}}, \ldots, \vz_{C}^{*\text{txt}} ] - [\vz_{1}^{\text{txt}}, \ldots, \vz_{C}^{\text{txt}} ]).
\end{split}
\end{equation}

We can also write any post-TTA text embedding $\vz_c^{\text{txt}}(\vtau^*)$ as a function of the text encoder $\mathbf{E}_{\text{txt}}$ prompted with optimized context vectors:
\begin{equation}
\begin{split}    
\vz_{c}^{\text{txt}}(\vtau^*) =  \mathbf{E}_{\text{txt}}([\vtau^*, \vt_c]) = \mathbf{E}_{\text{txt}}([\vtau - \lambda \nabla_{\vtau}H,\vt_c]).
\end{split}
\end{equation}

Through another first-order Taylor expansion (this time on $\vz_c^{\text{txt}}(\vtau^*)$), we have:
\begin{equation}
\begin{split}
   &\vz_{c}^{\text{txt}}(\vtau^*) = \mathbf{E}_{\text{txt}}([\vtau, \vt_c]) + (\nabla_{\vtau}  \mathbf{E}_{\text{txt}})^t (\vz_c^{\text{txt}}(\vtau^*) - \vz_c^{\text{txt}}(\vtau)) = \\
   &\mathbf{E}_{\text{txt}}([\vtau, \vt_c]) - \lambda (\nabla_{\vtau}  \mathbf{E}_{\text{txt}})^t  \nabla_{\vtau}(H),
\end{split}
\end{equation}
leading to an equivalent re-writing:
\begin{equation}\label{eq:zstarupdate}
\vz_{c}^{\text{txt}}(\vtau^*) = \vz_{c}^{\text{txt}}(\vtau) - \lambda (\nabla_{\vtau}  \mathbf{E}_{\text{txt}})^t  \nabla_{\vtau}(H)
\end{equation}

Substituting \eqref{eq:zstarupdate} into \eqref{eq:gexpanded}, we can express $g$ as follows:
\begin{equation}\label{eq:gexpandedgrad}
\begin{split}
 & g(c,\vz^{\text{img}}_i ,\vz_{1}^{*\text{txt}}, \ldots, \vz_{C}^{*\text{txt}}) = g(c,\vz^{\text{img}}_i ,\vz_{1}^{\text{txt}}, \ldots, \vz_{C}^{\text{txt}}) - \lambda  (\nabla_{[\vz_{1}^{\text{txt}}, \ldots, \vz_{C}^{\text{txt}} ]}g)^t\mathbf{d} \\
 &\text{where } \mathbf{d} \in \mathbb{R}^C \text{s.t.}~ \mathbf{d}_k = (\nabla_{\vtau} \mathbf{E}_{\text{txt}}([\vtau, \vt_k]))^t  \nabla_{\vtau} (H)([\vtau,\vt_k])) ~~\forall k \in \{1, \ldots, C\},
\end{split}
\end{equation}
with $\mathbf{d}_k$ denoting the $k$-th entry of the $C$ dimensional vector $\mathbf{d}$.
From \eqref{eq:gexpanded} and \eqref{eq:gexpandedgrad} the \emph{negative variation} $\delta g(c, \vz^{\text{img}})$ to $g$ before and after MEM can be expressed as:
\begin{equation}
\begin{split}
    &\delta g(c, \vz^{\text{img}}) = \lambda  (\nabla_{[\vz_{1}^{\text{txt}}, \ldots, \vz_{C}^{\text{txt}} ]}g)^t\mathbf{d}
\end{split}
\end{equation}

Finally, for any class, we can rewrite its final probability $p^{\text{end}}$ as a function of its initial probability $p^{\text{init}}$ and the variation of $g$ before and after TTA for the same class:

\begin{equation}\label{eq:final}
    p^{\text{end}}(c) = p^{\text{init}}(c) -\frac{\lambda}{N} \sum_i \delta g(c, \vz_i^{\text{img}})
\end{equation}

From Eq.\eqref{eq:final} we have that if $p^{\text{init}}(c^{\text{init}}) > \frac{\lambda}{N} \sum_i \delta g(c^{\text{init}}, \vz_i^{\text{img}})$, then the final probability $p^{\text{end}}(c^{\text{init}}) > 0$.
In the optimal case for MEM the entropy of $p^{\text{end}}$ is minimized, which entails that \emph{only one class} can have a probability strictly greater than 0.
Hence, $c^{\text{init}} = c^{\text{end}}$.

\end{proof}

\subsection{Experimental verification}
\input{tables/invariance}
We support the previous proposition with empirical evidence, by manually counting how often the prediction of $\overline{p}$ is invariant to Test-Time Prompt Tuning by MEM.
This experiment is easy to reproduce and consists of the following: augment $N$ times, filter by confidence, compute $p^{\text{init}}$, optimize by MEM, compute $p^{\text{end}}$ and check if $\argmax p^{\text{init}} = \argmax p^{\text{end}}$.
We report the proportion of samples for which the proposition holds for all Natural Distribution Shifts datasets in Table \ref{tab:invariance}, averaged over 3 runs with different seeds (the same used in Sec.~\ref{sec:exps} of the main body).
Although the proposition only accounts for the cases where entropy is globally minimized, the table shows that the marginal probability distribution is largely invariant to MEM. 
In the best case (ImageNet-Sketch) MEM alters the prediction of $\overline{p}$ only $8.77\%$ of the times. 
In the worst case (ImageNet-R), the prediction is unaltered for $96.78\%$ samples.

\subsection{Can invariance be anticipated?}
In the proof of Proposition 2.1, we express the post-MEM embeddings as a function of the pre-MEM embeddings through a Taylor expansion. 
For this relationship to hold, the variation needs to be small. 
If the initial entropy is high, the gradients from MEM (and, thus, the variation between pre- and post-MEM embeddings) can be larger than what a Taylor expansion can accurately approximate. In such cases, Prop. 2.1 cannot be guaranteed.
\input{tables/cr/app_ent_inv}
We execute a simple experiment using the validation set of ImageNet-1k, whose recipe is described below, to visualize this relationship. 

We compute pre- and post-MEM marginal probability distributions. We sort the pre-MEM distributions in order of descending entropy (most to least uncertain) and quantize them into 10 bins. 
Bins shall be interpreted as follows: the leftmost bin contains the top 10\% of samples with the highest entropy; the second bin contains samples outside the top-10\% percentile but within the top-20\%, and so on; the rightmost bin contains the bottom 10\% of samples with the lowest entropy.
For each bin we compute the invariance ratio, measuring how often the $\argmax$ of the pre-MEM $\overline{p}$ does \emph{not} change after MEM. 
Finally, we display a histogram with this data in Figure \ref{fig:entInv}.

A trend appears: as the entropy decreases (left to right), invariance holds more and more often. 
Hence, intuitively, the most likely cases where invariance to MEM does not hold are those of high uncertainty in the initial marginal probability distribution. 
However, this may still be rare: even within the top 10\% of most uncertain samples, invariance holds more than 82\% of the time (leftmost bin).
\clearpage

\section{Additional Experiments: LAION-2B Pretraining, Context Optimization and Hyperparameter Inheritance}\label{app:more_exp}
\input{tables/cr/app_nds_laion}
\input{tables/cr/app_coop}
\input{tables/cr/app_fg_laion}
This Appendix deals with enriching the experiments of Section \ref{sec:exps}, which focused on models officially released by OpenAI \cite{openai2024clip}.
Here we focus on the comparison with TPT \cite{shu2022test} and extend the analysis to: \ding{172} \texttt{CLIP-ViT-B-16} pretrained on the 2B English Subset of LAION-5B \cite{schuhmann2022laion}; \ding{173} OpenAI's CLIP, transferred after supervised Context Optimization (\texttt{CoOp}) \cite{zhou2022learning}. 

\textbf{Implementation Details.}
For the experiments with LAION Pretraining, we use the \texttt{open\_clip} repository, \ie, the official code for \cite{cherti2023reproducible}.
The pretrained keyword for this model is \texttt{laion2b\_s34b\_b88k}.
For \texttt{CoOp} we use the context vectors learned on ImageNet-1k officially released by \cite{zhou2022learning}. 
The experimental setup is analogous to Section \ref{sec:exps} in all details.
We do not tune any hyperparameters for these different initializations, but inherit them from the experiments with OpenAI models.

\subsection{LAION-2B Pretraining}
Table \ref{tab:nds_laion} reports experiments on Natural Distribution Shifts, from which we observe no differences \wrt OpenAI models: \method largely outperforms TPT, and peak difference is reached with ImageNet-A \cite{hendrycks2021natural}.
Results on Fine-grained Classification are given in Table \ref{tab:fc_laion}.
We observe that \method improves the zero-shot baseline better with this pretraining, and overcomes TPT with an average margin of $+0.4\%$.
In contrast, ensembling textual prompts appears less effective. 
We speculate this is because the 7 templates were explicitly tuned and selected for OpenAI models. 
The worst-case scenario is confirmed with satellite imagery \cite{helber2019eurosat}; please refer to Appendix \ref{app:eurosat} for a deeper investigation.

\subsection{Context Optimization (CoOp)}
For this comparison, we follow \cite{shu2022test} and report \texttt{CoOp} on Natural Distribution Shifts only, presenting results in Table \ref{tab:nds_coop}.
We further observe patterns consistent with OpenAI models, with \method providing large improvements over TPT.
Also here, the best-case scenario persists with ImageNet-A. 

\subsection{Hyperparameter Inheritance}
\input{tables/robust}
\input{tables/finegrained}
In all experiments so far, including Section \ref{sec:exps} as well as Tables \ref{tab:nds_laion}, \ref{tab:nds_coop} and \ref{tab:fc_laion}, we employed a percentile for confidence-based filtering set to 0.3. 
This value was obtained after validation on ImageNet-1k with OpenAI's \texttt{CLIP-ViT-B-16} and kept fixed for all models and datasets.
Here, we show that \method obtains favorable performance even if the percentile for confidence-based filtering is not tuned in any way, but set to 0.1 by ``inheriting'' the value used in TPT \cite{shu2022test}. 
These results are given in Tables \ref{tab:nds-inherited} and \ref{tab:fc-inherited}.
Surprisingly, some datasets within the Natural Distribution Shifts benchmark benefit from this more restrictive filtering (ImageNet-A above all), while we observe that Finegrained classification tends to improve when more views are retained. 
The core findings, however, are entirely unchanged: the best case remains ImageNet-A, the worst-case remains EuroSAT, and \method outperforms competitor in most datasets, no matter the experimental setup.

\section{Calibration and Overconfidence of CLIP on augmented Natural Distribution Shifts}\label{app:overconfidence}
\input{tables/ece}
In Section \ref{sec:overconfidence} of the manuscript, the validation set of ImageNet-1k is shown to convey that overconfidence emerges as a critical issue when predicting over augmented views.
In this appendix, we expand the analysis to the 4 datasets for robustness to Natural Distribution Shifts (NDS) \cite{hendrycks2021natural, hendrycks2021many, recht2019imagenet, wang2019learning}.
For all datasets, we follow the augmentation setup of Sec.\ref{sec:overconfidence}, and generate augmented counterparts with $6\times$ more examples.

First, let us define the calibration of DNNs. Calibrating DNNs is crucial for developing reliable and robust AI systems, especially in safety-critical applications. 
A DNN is perfectly calibrated if the probability that its prediction is correct ($\hat{y}= y$) given a confidence score random variable $S$ is equal to its confidence score.
The confidence score is commonly taken as the maximum of the output probability vector of the model, \ie, $s=\max p(\cdot)$:
\begin{equation}
 P(\hat{y}= y|S=s)=s
\end{equation}

To evaluate the expected calibration error (ECE), we typically split the dataset into $M$ bins $B_m$ based on their confidence scores. We then calculate the accuracy of each bin, denoted as $\text{acc}(B_m)$, and the average confidence, denoted as $\text{conf}(B_m)$. The ECE is defined by the following formula:

\begin{equation}
\text{ECE} = \frac{1}{M} \sum_m^M \| \text{acc}(B_m) - \text{conf}(B_m)\|
\end{equation}

Then, we show how the ECE of \texttt{CLIP-ViT-B-16} varies between ``source'' and augmented versions of all datasets (ImageNet-1k included) in Figure \ref{fig:ece}.
From this experiment, we observe a large increase in the ECE across all datasets.
In no cases, the ECE remains comparable to its default value when no augmentations are present.
As we discussed in \ref{sec:overconfidence}, the calibration error increases when the model is either more accurate than confident (signaling \emph{underconfidence}) or the opposite, signaling \emph{overconfidence}. 
Reliability diagrams are a standard tool to understand which is the case, hence we show them for all 4 NDS Datasets in Fig.\ref{fig:rpAll}.
\input{tables/rpAll}
\input{tables/cr/rp_laiob2b}
These results are entirely consistent with Sec.\ref{sec:overconfidence}: the calibration error increases exclusively due to overconfidence, no matter the dataset. 
In parallel, the error rate of \texttt{CLIP-ViT-B-16} can either remain close to its default value (\eg, ImageNet-Sketch), slightly decrease (\eg, ImageNet-R and -v2) or largely decrease (ImageNet-A).
We observe an identical pattern for CLIP models pretrained on LAION. 
For reference, see Figure \ref{fig:rp_laion}.

\section{On the expected risk of $\overline{p}$ and $p$.}\label{app:risk}
The \emph{expected} risk of a classifier $f$ is commonly defined as the expectation of the risk function $\ell$ over the joint distribution of data and labels.
\begin{equation}
  \mathcal{R}(f) =\mathbb{E}_{(x,y)\sim P_{\mathcal{X}\mathcal{Y}}}\Big[\ell(y, f(\vx))\Big].
\end{equation}
In \cite{kimura2021understanding}, the expected risk of a classifier $\overline{f}(\vx) = \overline{p}(\cdot|\vx)$, which predicts by marginalizing over several augmented views, is theoretically shown to lower-bound the empirical risk of a standard classifier $f=p(\cdot|\vx)$ when the risk function $\ell$ is a squared error, \ie, $\ell(a,b)=(a-b)^2$.

Here, we show that such a bound can be extended to any risk function $\ell$ that checks the triangular inequality.
Specifically, note that if $\ell$ satisfies the triangular inequality, then: 
\begin{equation}\label{eq:triangle}
    \ell(y, \overline{p}(\vx)) \leq \frac{1}{N}\sum_{i=1}^N\ell(y, p(\vx_i)).
\end{equation}
The above inequality is obtained following these simple steps:
\begin{equation}
\| y- \overline{p}(\vx) \| = \|y - \frac{1}{N} \sum_{i=1}^N p(\vx_i)\|  = \| \frac{1}{N} \sum_{i=1}^N (y -  p(\vx_i)) \|\leq \frac{1}{N}\sum_{i=1}^N  \| (y -  p(\vx_i)) \|
\end{equation}

Applying the expectation operator $\mathbb{E}$ over the joint distribution ${P}_{\mathcal{X}\mathcal{Y}}$ to both sides of Eq.\eqref{eq:triangle} leads to:
\begin{equation}
  \mathcal{R}(\overline{p}) \leq  \frac{1}{N}\sum_{i=1}^N \mathcal{R}(p)=  \mathcal{R}(p).
\end{equation}
Hence, the empirical risk of $\overline{p}$ lower-bounds that of $p$ for any risk function $\ell$ satisfying the triangular inequality.

\section{Tie breaking with \textsc{Zero}}
A caveat of \method are ties, \ie, cases with multiple classes having identical probability within the marginal probability distribution.
This is clear to see when viewing \method as its equivalent paradigm of voting among confident views, simply because more than one class may have an equal amount of ``votes''.
Throughout all the experiments of this work, ties are broken \emph{greedily}.
If a tie results from the top views, the procedure for breaking it follows these two steps: \ding{172} sort the remaining views by ascending entropy (most to least confident) and \ding{173} scan the views until a prediction is encountered that breaks the tie.
Other than this, many alternative are possible, such as relying on the most confident prediction.
Specifically, we have explored the following alternatives:
\begin{enumerate}
[leftmargin=2em]
    \item greedy tie breaking, as discussed above;
    \item relying on the most confident prediction; 
    \item computing several marginal probabilities for $\overline{p}$, each by marginalizing over views with identical predictions, and picking the one with the lowest entropy for the final decision;
    \item relying on the maximum logit (pre-Softmax);
    \item using the averaged logits (pre-Softmax);
    \item doing similar to point 2, using logits instead of probabilities;
    \item random tie breaking;
\end{enumerate}
and did not find consistent behaviour across all (fine-grained and NDS) datasets, suggesting this is indeed a minor component.
We opted for greedy tie breaking due to its slightly better performance on the ImageNet validation set.

\section{A failure mode for TTA: satellite imagery}\label{app:eurosat} 
In our experiments, we find that an extremely OOD domain represents a consistent failure mode for TTA: satellite imagery. 
In all comparison groups, a zero-shot baseline largely outperforms \emph{any} TTA strategy when evaluated on EuroSAT\cite{helber2019eurosat}: 
\begin{itemize}
[leftmargin=2em]
    \item in \emph{Group 1} the zero-shot baseline \texttt{CLIP-Ensemble} largely outperforms the best TTA strategy \textsc{Zero}\textsubscript{+Ensemble};
    \item in \emph{Group 2}, zero-shot \texttt{MaPLe} outperforms PromptAlign; 
    \item in \emph{Group 3} the best RLCF$_{t=1}^{\Theta_v}$ pipeline lies far behind the zero-shot teacher \texttt{CLIP-ViT-L-14}.
\end{itemize} 
Here, we qualitatively and quantitatively report two main root causes for failures. %

\textbf{Qualitatively poor augmentations.} In principle, TTA methods should rely on generic data augmentations, since not doing so would require going against the principles of the field by assuming some prior knowledge about the test data is available.
As discussed in Sec.\ref{sec:baseline}, data augmentations are a doubled edged sword in TTA, and failing in crafting properly augmented views can potentially
generate misleading or uninformative visual signals.
We report some qualitative examples conveying this problem in Figure \ref{fig:esat}.
In the Figure, we report three images from \cite{helber2019eurosat}, together with the top-3 augmented views leading \texttt{CLIP-ViT-B-16} to its most confident predictions.
Each source image is reported with the groundtruth, and all views are reported with both the prediction and the confidence of CLIP.
Visually, one can perceive that the simple data augmentation scheme of cropping and flipping, which has largely been proven successful in \cite{shu2022test, samadh2023align} and in our work, does not provide informative views, since most are alike one another.

\textbf{Quantitatively high error.} Augmentations are used by all TTA methods discussed in this paper, hence the previous discussion holds for TPT as much as it does for PromptAlign, RLCF or \method. \input{tables/rpEsat} 
Nevertheless, we highlight an additional caveat about satellite imagery which is particularly detrimental for \method, and relates to the base model error over augmentations.
Recall that, in \method, the usage of $\overline{p}$ is backed by theoretical motivations, and the manual adaptation of the temperature is supported by two concurrent observations: augmentations-induced overconfidence and a comparable error rate between source and augmented images.
Simply put, the latter condition is not verified for satellite imagery.
To show this phenomenon, we follow the experimental setup of Sec.\ref{sec:overconfidence} and examine the reliability diagrams of EuroSAT\cite{helber2019eurosat} and of its augmented counterpart in Figure \ref{fig:rpEsat}.
As per Section \ref{sec:overconfidence}, we display the ECE and the Top1-Accuracy on each version of the dataset.
From this perspective, one can note that the base model error largely increases, in this domain, when augmented views are present. 
The accuracy on source images is $42.01\%$, dropping to $35.21\%$ simply due to augmentations.

Both observations, combined, suggest that crafting augmentations for satellite imagery requires an ad-hoc treatment, which makes it a controversial benchmark for TTA.

\section{Natural Distribution Shifts vs Fine-grained Classification}
Throughout the manuscript, one can observe that \method consistently provides larger improvements in Natural Distribution Shifts than it does in the Finegrained suite.
We thus devote this section to digging deeper into this matter.

Perhaps unsurprisingly, we posit that \method improves over the zero-shot baseline if the zero-shot error rate of the model does not largely increase with augmented views. 
As Fig.\ref{fig:preliminary}(b) displays, this is the case for all Natural Distribution Shifts datasets. 
To understand any different behaviors, we repeat the same experiment of Section \ref{sec:overconfidence} for the entire Fine-grained suite and report the results in Table \ref{tab:error_gap}.
\input{tables/cr/app_aug_error}
Please note that, in the table, the percentile for confidence-based in filtering in \method is set to $0.1$, since the protocol for generating the augmented datasets follows the setup of TPT, which also uses a cutoff percentile of $0.1$, and that we omit EuroSAT since an analogous experiment was presented in the previous Appendix.

Overall, we observe a strong correlation between the error gap and the improvement provided by Zero, with Spearman’s coefficient being $-0.95$ across datasets.
This result shows that the correlation is negative, \ie, the lower the error gap, the larger the improvement (or, in other words, the better the zero-shot performance on augmented views, the larger the improvement of \method).
This pattern is also consistent with the experiments on EuroSAT reported in the previous Appendix.
Understanding why augmentations induce larger or smaller errors may be a case-by-case matter that relates to the nature of the datasets. 
Here, we pinpoint two possible reasons:
\begin{itemize}
[leftmargin=2em]
    \item The semantic space of the ImageNet variants of the Natural Distribution Shifts benchmark comprises many common categories, which may have appeared frequently during CLIP’s pretraining. 
    Hence, it seems reasonable that CLIP is robust \emph{w.r.t.} augmented views of images belonging to these categories. 
    In the Fine-grained classification suite, datasets such as SUN397 and Caltech101 also contain common object categories, which is consistent with the results shown above. 
    In contrast, other datasets such as Flowers102 and Oxford-Pets span much less frequent concepts.
    \item Other than the semantic classification space, images' visual appearance also plays an important role. 
    For example, datasets such as FGVC-Aircraft and Stanford Cars still contain rare concepts, but \method largely improves over the baseline nonetheless. 
    Our augmentation setup is simple, and only contains random resized crops and random horizontal flips, which can constitute a “zoom-in” to a random portion of the image. 
    For some benchmarks, this is useful as it may trigger CLIP’s capabilities to recognize small details, such as logos, or even reading text, such as the car brand or the airline name.
    In contrast, more object-centric datasets such as Flower102, may lead to missing precious visual features (\eg, the stem).
\end{itemize}

In our work we did not search for the best data augmentations but rather stuck to an established setting, using the same augmentations setup for all datasets. 
Nevertheless, the performance of \method is linked to the impact that data augmentations have on how the model perceives images, and we believe this is an interesting research direction to pursue.

\section{Independence among views in the setup of Test-Time Adaptation}\label{app:independence}
The theoretical framework of Section \ref{sec:margBound} models an ideal scenario, where independence holds among different inputs. 
To clarify, this means that the model's error on view $\vx_i$ should not be correlated with the error on any other view $\vx_j$, which allows writing the compound error with a binomial distribution as in \eqref{eq:majErr}.

In practice, achieving perfect independence is challenging, if not impossible. 
Hence, a suitable approximation strategy to mitigate this issue is to promote diversity. 
In classical ensembling theory, a well-established approach is to train different models on different subsets of the available data.
Similarly, the augmentation scheme of random cropping aligns with this approach by presenting the model with different portions of the image each time.

Moreover, ideally, the augmentation pipeline should not change the underlying label of the original input and guarantee that the model’s error rate on augmented views remains comparable to the error rate on the original inputs belonging to the same category. 
In practice, this entails that augmentations should not disrupt the visual appearance of the image, and, consequently, some views may result in a slight or moderate correlation, because some “parts” of the source image will overlap among them.
An analogy with classical literature can be drawn also in this case.
Specifically, when not enough data are available, overlaps among the training sets of different models are required to ensure convergence.
Consequently, models producing slightly or moderately correlated predictions are more likely to emerge.

\section{Additional Implementation Details}
\input{tables/cr/app_std}
\textbf{Standard deviations.} To complement the results on Fine-grained classification, we report the standard deviation of \method computed over 3 runs with different seeds in Table \ref{tab:std}.
These are not reported together with the average top1-accuracy in Tab.~\ref{tab:fc} to avoid an excessively dense table.
On average, standard deviations are very small, suggesting that regardless of the inherent randomness of data augmentations, \method is relatively stable.
Note that standard deviations in \emph{Group 2} (\ie, with \texttt{MaPLe}) are slightly greater than those in the other groups. 
This fact does not stem from \method's or \texttt{MaPLe}'s greater instability, but from an experimental detail which we report here for completeness: while only one set of weights is officially released for each CLIP version \cite{openai2024clip}, \citet{khattak2023maple} released 3 sets of pretrained weights for \texttt{MaPLe}, varying on the seed.
To avoid picking one, we associated a set of weights to each of our runs, hence results from slightly different initializations are computed to match the experimental setup of \citet{samadh2023align} (PromptAlign).

\textbf{Reproducibility of TTA methods.} For section \ref{sec:exps}, we reproduced all methods using the source code provided by the authors with the hardware at our disposal.
This was done to ensure that hardware differences did not interfere with a correct evaluation.
We found that all TTA strategies are highly reproducible, with negligible differences (\ie, $\Delta <0.1$) which we omitted by reporting the numbers from the official papers.
In case of larger differences, we reported reproduced results.

\input{tables/esatImages}

%% file: tables/invariance.tex
\begin{table}[h]
\def\arraystretch{1.1}
    \caption{Empirical evidence supporting Proposition 2.1.}
    \centering
    \scriptsize 
    \begin{tabularx}{\textwidth}{
    >{\centering\arraybackslash}p{4cm}
     >{\centering\arraybackslash}X
     >{\centering\arraybackslash}X
     >{\centering\arraybackslash}X
     >{\centering\arraybackslash}X
     >{\centering\arraybackslash}X
    }
    \toprule

    \textbf{Proposition} & IN-1k & IN-A & IN-v2 & IN-R & IN-Sketch \\
    \cmidrule(lr){1-6}

    $\argmax p^{\text{init}} = \argmax p^{\text{end}}~[\%]$ & 95.73$\pm0.05$ & 95.55$\pm0.12$ & 94.86$\pm0.17$ & 96.78$\pm0.08$ & 91.23$\pm0.09$\\
    \bottomrule

    \end{tabularx}
    \label{tab:invariance}
\end{table}

%% file: tables/cr/app_ent_inv.tex
\begin{figure}[b!]
    \centering
    \includegraphics[width=0.65\textwidth]{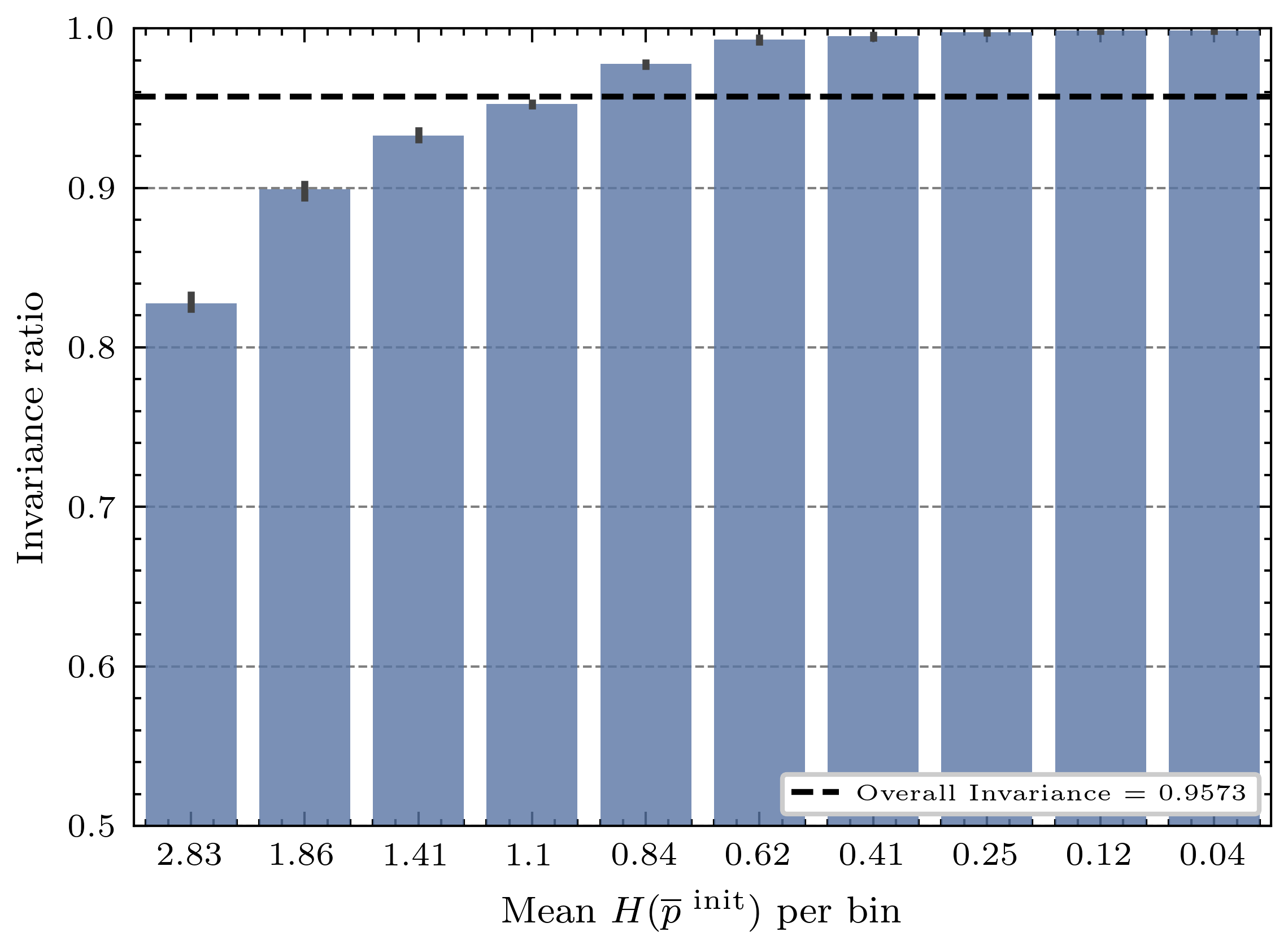}
    \caption{Entropy of the pre-TTA marginal probability distribution vs the invariance ratio.}
    \label{fig:entInv}
    \vspace{-1em}
\end{figure}

%% file: tables/cr/app_nds_laion.tex
\begin{table}[t!]
    \def\arraystretch{1.15}
    \caption{Results on Natural Distribution Shifts when adapting \texttt{CLIP-ViT-B-16} pretrained on the 2B English subset of LAION-5B. Top-1 accuracy is reported, and \textbf{bold text} indicates the best performer.}
    \centering
    \scriptsize
    {\begin{tabularx}{\textwidth}{
     >{\raggedright\arraybackslash}X
     >{\centering\arraybackslash}X
     >{\centering\arraybackslash}X
     >{\centering\arraybackslash}X
     >{\centering\arraybackslash}X
     >{\centering\arraybackslash}X
     >{\centering\arraybackslash}X
    }
     \toprule
    \textbf{Method} & ImageNet & A & V2 & R & Sketch & Mean \\

    \midrule
    \rowcolor{lightgray}
    \multicolumn{7}{c}{\texttt{CLIP-ViT-B-16 (LAION2B)}} \\
    [-2ex] \\
    \emph{Zero-Shot} & 69.27 & 37.08 & 61.27 & 78.83 & 54.85 & 60.26\\
    
    \emph{Ensemble} &  70.43 & 38.32 & 62.28 & 80.41 & 55.54 & 61.40 \\
    
    TPT & 70.61 & 41.94 & 62.96 & 80.40 & 55.48 & 62.28 \\
    
    \rowcolor{zerocolor}
    \textsc{Zero} & 71.39 & 48.71 & 63.53 & 80.59 & 55.82 & 64.01 \\
    
    \rowcolor{zerocolor}
    \textsc{Zero}\textsubscript{+Ensemble} & \textbf{72.14} & \textbf{49.02} & \textbf{64.32} & \textbf{82.42} & \textbf{56.53} & \textbf{64.89} \\

     \bottomrule
    \end{tabularx}}
    \label{tab:nds_laion}
\end{table}

%% file: tables/cr/app_coop.tex
\begin{table}[t!]
    \def\arraystretch{1.075}
    \caption{Results on Natural Distribution Shifts when adapting OpenAI's \texttt{CLIP-ViT-B-16}, with \texttt{CoOp}-learned prompts. Top-1 accuracy is reported, and \textbf{bold text} indicates the best performer.}
    \centering
    \scriptsize
    \begin{tabularx}{\textwidth}{
     >{\raggedright\arraybackslash}p{1.8cm}
     >{\centering\arraybackslash}X
     >{\centering\arraybackslash}X
     >{\centering\arraybackslash}X
     >{\centering\arraybackslash}X
     >{\centering\arraybackslash}X
     >{\centering\arraybackslash}X
    }
     \toprule
    \textbf{Method} & ImageNet & A & V2 & R & Sketch & Mean \\

    \midrule
    \rowcolor{lightgray}
    \multicolumn{7}{c}{\texttt{CoOp}} \\
    [-2ex] \\
    \emph{Zero-Shot} & 71.51 & 49.71 & 64.20 & 75.21 & 47.99 & 61.72 \\
    TPT & 73.64 & 57.77 & 66.72 & 78.03 & 49.56 & 65.14 \\

    \rowcolor{zerocolor}
    \textsc{Zero} & \textbf{74.12} & \textbf{61.57} & \textbf{67.15} & \textbf{78.43} & \textbf{49.77} & \textbf{66.21} \\

     \bottomrule
    \end{tabularx}
    \label{tab:nds_coop}
\end{table}

%% file: tables/cr/app_fg_laion.tex
\begin{table}[t!]
    \caption{Fine-grained Classification with \texttt{CLIP-ViT-B-16} pretrained on the 2B English subset of LAION-5B. Top-1 accuracy is reported, and \textbf{bold text} indicates the best performer.}
    \def\arraystretch{1.15}
    \centering
    \scriptsize
    \begin{tabularx}{\textwidth}{
    >{\raggedright\arraybackslash}p{1.5cm}
    >{\centering\arraybackslash}X
    >{\centering\arraybackslash}X
    >{\centering\arraybackslash}X
    >{\centering\arraybackslash}X
    >{\centering\arraybackslash}X
    >{\centering\arraybackslash}X
    >{\centering\arraybackslash}X
    >{\centering\arraybackslash}X
    >{\centering\arraybackslash}X
    >{\centering\arraybackslash}X
    >{\centering\arraybackslash}X
    >{\centering\arraybackslash}X
    }
    
    \toprule
    \textbf{Method} & FLWR & DTD & PETS & CARS & UCF & CAL & FOOD & SUN & AIR & ESAT & Mean & Median \\

    \midrule

    \rowcolor{lightgray}
    \multicolumn{13}{c}{\texttt{CLIP-ViT-B-16 (LAION2B)}} \\ [-2ex] \\ 
     
    \emph{Zero-Shot} & 69.71 & 54.43 & 89.37 & 89.94 & 64.02 & 95.82 & 81.38 & 70.60 & 26.04 & 47.05 & 68.84 & 70.16 \\

    \emph{Ensemble} & 68.70 & 54.55 & 87.76 & 89.98 & 67.64 & 96.51 & 81.64 & 70.62 & 25.68 & \textbf{49.64} & 69.27 & 69.66\\
    
    TPT & 69.47 & 54.53 & 89.00 & 90.72 & 66.68 & 96.16 & 81.76 & \textbf{71.34} & 26.73 & 48.81 & 69.52 & 70.41 \\

    \rowcolor{zerocolor}
    \textsc{Zero} & \textbf{70.82} & 55.20 & \textbf{89.77} & \textbf{91.95} & 67.23 & 96.13 & \textbf{83.65} & 71.21 & \textbf{28.25} & 45.01 & 69.92 & \textbf{71.02} \\

    \rowcolor{zerocolor}
    \textsc{Zero}\textsubscript{+Ensemble} & 68.01 & \textbf{55.95} & 87.67 & 91.87 & \textbf{69.11} & \textbf{96.54} & 83.83 & 71.09 & 28.10 & 47.10 & \textbf{69.93} & 70.10 \\
    
    \bottomrule
    \end{tabularx}

    \label{tab:fc_laion}
\end{table}

%% file: tables/robust.tex
\begin{table}[t!]
    \def\arraystretch{1.075}
    \caption{Natural Distribution Shifts (percentile = 0.1). TTA methods are grouped according to the baseline model and top-1 accuracy is reported. \textbf{Bold text} is the best method within each group.}
    \centering
    \scriptsize{\begin{tabularx}{\textwidth}{
     >{\raggedright\arraybackslash}X
     >{\centering\arraybackslash}X
     >{\centering\arraybackslash}X
     >{\centering\arraybackslash}X
     >{\centering\arraybackslash}X
     >{\centering\arraybackslash}X
     >{\centering\arraybackslash}X
    }
     \toprule
    \textbf{Method} & \textbf{ImageNet} & \textbf{A} & \textbf{V2} & \textbf{R} & \textbf{Sketch} & \textbf{Average} \\

    \midrule
    \rowcolor{lightgray}
    \multicolumn{7}{c}{\texttt{CLIP-ViT-B-16}} \\
    [-2ex] \\
    Zero-Shot & 66.73 & 47.87 & 60.86 & 73.98 & 46.09 & 59.11 \\
    Ensemble & 68.34 & 49.89 & 61.88 & 77.65 & 48.24 & 61.20 \\
    TPT & 68.98 & 54.77 & 63.45 & 77.06 & 47.94 & 62.44 \\
    
    \rowcolor{zerocolor}
    \textsc{Zero} &  {69.06}{\scriptsize$\pm$0.04} &  {61.35}{\scriptsize$\pm$0.26} &  {64.13}{\scriptsize$\pm$0.17} &  {77.28}{\scriptsize$\pm$0.08} &  {48.29}{\scriptsize$\pm$0.04} &  {64.02} \\
    
    \rowcolor{zerocolor}
    \textsc{Zero}\textsubscript{+Ensemble} & \textbf{70.93}{\scriptsize$\pm$0.02} & \textbf{64.06}{\scriptsize$\pm$0.09} & \textbf{65.16}{\scriptsize$\pm$0.21} & \textbf{80.75}{\scriptsize$\pm$0.08} & \textbf{50.32}{\scriptsize$\pm$0.09} & \textbf{66.24} \\

    \cmidrule(lr){1-7}
    \rowcolor{lightgray}
    \multicolumn{7}{c}{\texttt{MaPLe}} \\
    [-2ex] \\
    Zero-Shot & - & 50.90 & 64.07 & 76.98 & 49.15 & 60.28 \\
    TPT & - & 58.08 & 64.87 & 78.12 & 48.16 & 62.31 \\
    PromptAlign & - & 59.37 & 65.29 & 79.33 & 50.23 & 63.55 \\
    
    \rowcolor{zerocolor}
    \textsc{Zero} & - & \textbf{64.65}{\scriptsize$\pm$0.24} & \textbf{66.63}{\scriptsize$\pm$0.32} & \textbf{79.75}{\scriptsize$\pm$0.41} & \textbf{50.73}{\scriptsize$\pm$0.62} & \textbf{65.44} \\

    \cmidrule(lr){1-7}
    \rowcolor{lightgray}
    \multicolumn{7}{c}{\texttt{CLIP-ViT-B-16 + CLIP-ViT-L-14}} \\
    [-2ex] \\
    ZeroShot & 73.44 & 68.82 & 67.80 & 85.40 & 57.84 & 70.66 \\  
    RLCF~$_{t=3}^{\vtau}$ & 73.23 & 65.45 & 69.77 & 83.35 & 54.74 & 69.31 \\ [-2.5ex] \\
    RLCF~$_{t=3}^{\Theta_v}$ & \textbf{74.85} & 73.71 & \textbf{69.77} & 86.19 & 57.10 & 72.32 \\ [-2.5ex] \\
    
    \rowcolor{zerocolor}
    \textsc{Zero} & 74.48{\scriptsize$\pm$0.12} & \textbf{77.07}{\scriptsize$\pm$0.35} & 69.53{\scriptsize$\pm$0.12} & \textbf{86.87}{\scriptsize$\pm$0.05} & \textbf{58.59}{\scriptsize$\pm$0.08} & \textbf{73.31} \\

     \bottomrule
    \end{tabularx}}
    \label{tab:nds-inherited}
\end{table}

%% file: tables/finegrained.tex
\begin{table}[t!]

    \def\arraystretch{1.15}
    \centering
    \caption{Finegrained classification (percentile = 0.1). Formatting follows other tables.}
    \scriptsize
    \begin{tabularx}{\textwidth}{
    >{\raggedright\arraybackslash}p{1.5cm}
    >{\centering\arraybackslash}X
    >{\centering\arraybackslash}X
    >{\centering\arraybackslash}X
    >{\centering\arraybackslash}X
    >{\centering\arraybackslash}X
    >{\centering\arraybackslash}X
    >{\centering\arraybackslash}X
    >{\centering\arraybackslash}X
    >{\centering\arraybackslash}X
    >{\centering\arraybackslash}X
    >{\centering\arraybackslash}X
    >{\centering\arraybackslash}X
    }
    
    \toprule
    \textbf{Method} & FLWR & DTD & PETS & CARS & UCF & CAL & FOOD & SUN & AIR & ESAT & Mean & Median \\

    \midrule
    \rowcolor{lightgray}
    \multicolumn{13}{c}{\texttt{CLIP-ViT-B-16}} \\ [-2.25ex] \\
     
    Zero-Shot &  {67.44} & 44.27 &  {88.25} & 65.48 & 65.13 & 93.35 & 83.65 & 62.59 & 23.67 & 42.01 & 63.58 & 65.31 \\
     
    Ensemble & 67.07 & 45.09 & \textbf{88.28} & 66.16 & 67.51 & 93.91 & 84.04 &  {66.26} & 23.22 & \textbf{50.42} & \textbf{65.20} & 66.66 \\

    TPT & \textbf{68.75} & \textbf{47.04} & 87.23 & 66.68 &  {68.16} &  {93.93} & \textbf{84.67} & 65.39 & 23.13 & 42.86 & 64.78 & 67.42 \\

    \rowcolor{zerocolor}
    \textsc{Zero} & 67.07 & 45.80 & 86.74 &  {67.54} & 67.64 & 93.51 & 84.36 & 64.49 &  {24.40} & 39.60 & 64.11 & 67.31 \\

    \rowcolor{zerocolor}
    \textsc{Zero}\textsubscript{+Ensemble} & 66.82 & 45.86 & 87.20 & \textbf{68.48} & \textbf{68.57} & \textbf{94.14} & 84.58 & \textbf{66.90} & \textbf{24.42} & 43.77 & 65.07 & \textbf{67.69} \\

    \cmidrule(lr){1-13}
    \rowcolor{lightgray}
    \multicolumn{13}{c}{\texttt{MaPLe}} \\  [-2.25ex] \\
    Zero-Shot & 72.23 & 46.49 & 90.49 & 65.57 & 68.69 & 93.53 & 86.20 & 67.01 & 24.74 & 48.06 & 66.30 & 67.85 \\  
    TPT & 72.37 & 45.87 & 90.72 & 66.50 & 69.19 & 93.59 & 86.64 & 67.54 & 24.70 & 47.80 & 66.49 & 68.37 \\
    PromptAlign & \textbf{72.39} & 47.24 & \textbf{90.76} & \textbf{68.50} & 69.47 & 94.01 & 86.65 & 67.54 & 24.80 & 47.86 & \textbf{66.92} & \textbf{68.98} \\  
    
    \rowcolor{zerocolor}
    \textsc{Zero} & 71.20 & \textbf{47.70} & 90.17 & 67.91 & \textbf{69.49} & \textbf{94.12} & \textbf{86.78} & \textbf{67.55} & \textbf{25.57} & 41.05 & 66.15 & 68.70 \\

    \cmidrule(lr){1-13}
    \rowcolor{lightgray}
    \multicolumn{13}{c}{\texttt{CLIP-ViT-B-16 + CLIP-ViT-L-14}} \\ [-2.25ex] \\
     
    ZeroShot & \textbf{75.76} & 51.83 &  {92.86} &  {76.16} &  {73.70} & 94.04 & \textbf{88.03} & 66.96 & 30.54 & \textbf{54.38} & \textbf{70.43} & 74.73 \\
     
    RLCF~$_{t=1}^{\vtau}$ & 71.58 & 50.34 & 89.01 & 69.76 & 69.84 & 94.09 & 85.90 & 67.33 & 23.71 & 46.87 & 66.84 & 69.80 \\  
    RLCF~$_{t=3}^{\vtau}$ & 72.49 & 51.93 & 89.55 & 72.91 & 72.31 & \textbf{95.00} & 86.84 &  {69.04} & 25.40 & 45.96 & 68.14 & 72.40 \\  
    RLCF~$_{t=1}^{\Theta_v}$ & 72.56 & 52.21 & 89.51 & 63.12 & 71.49 & 94.65 & 86.90 & 68.50 & 24.06 & 47.74 & 67.07 & 70.00 \\  
    RLCF~$_{t=3}^{\Theta_v}$ & 71.74 & {53.27} & 91.15 & 70.93 & 73.24 &  {94.73} & 87.28 & \textbf{69.38} & 28.54 & 47.41 & 68.77 & 71.34 \\  
    
    \rowcolor{zerocolor}
    \textsc{Zero} &  {75.34} & \textbf{54.22} & \textbf{92.90} & \textbf{77.33} & \textbf{74.26} & 94.52 &  {87.57} & 68.05 & \textbf{32.11} & 42.74 & 69.90 & \textbf{74.80} \\
    
    \bottomrule
    \end{tabularx}

    \label{tab:fc-inherited}
\end{table}

%% file: tables/ece.tex
\begin{wrapfigure}{o}{0.48\textwidth}
    \centering
    \vspace{-1.5em}
    \includegraphics{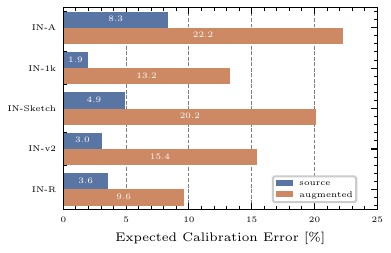}
    \caption{Expected Calibration Error (ECE) \cite{guo2017calibration} of \texttt{CLIP-ViT-B-16} across 5 datasets for robustness to natural distribution shifts. Blue is the ECE of zero-shot \texttt{CLIP}, and orange is the ECE of zero-shot \texttt{CLIP} on an augmented version of the dataset after confidence-based thresholding.}
    \label{fig:ece}
\end{wrapfigure}

%% file: tables/rpAll.tex
\begin{figure}[t!]
    \setlength{\abovecaptionskip}{2pt}
    \centering
    \includegraphics{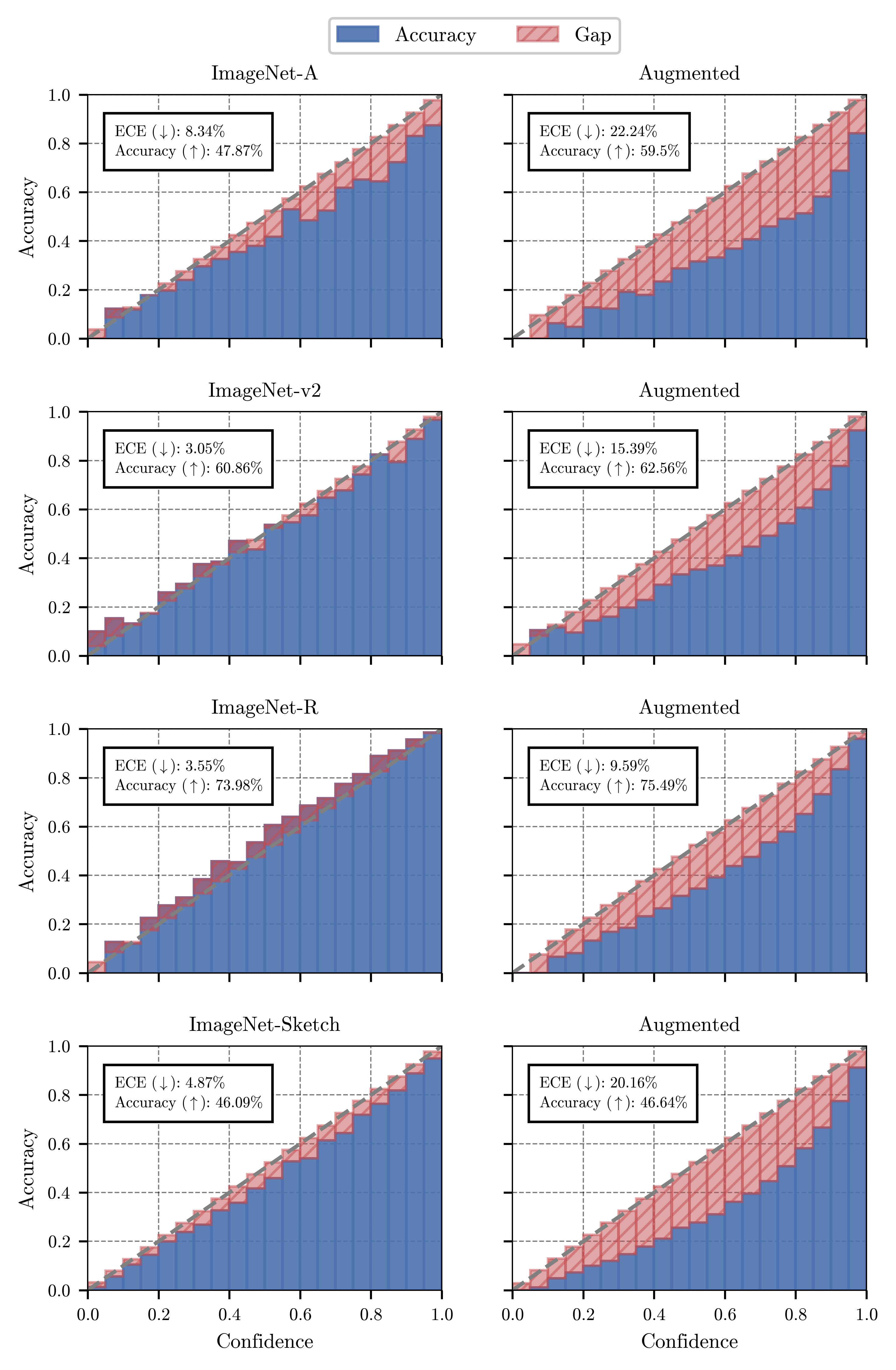}
    \caption{Reliability diagrams (20 bins) for \texttt{CLIP-ViT-B-16} on the 4 datasets for Natural Distribution Shifts. 
    In each row, left is the ECE on the source dataset, right on the augmented and filtered version. Row 1: ImageNet-A \cite{hendrycks2021natural}; Row 2: ImageNet-v2 \cite{recht2019imagenet}; Row 3: ImageNet-R \cite{hendrycks2021many}; Row 4: ImageNet-Sketch \cite{wang2019learning}.}
    \label{fig:rpAll}
\end{figure}

%% file: tables/cr/rp_laiob2b.tex
\begin{figure}
    \centering
    \includegraphics[width=0.8\linewidth]{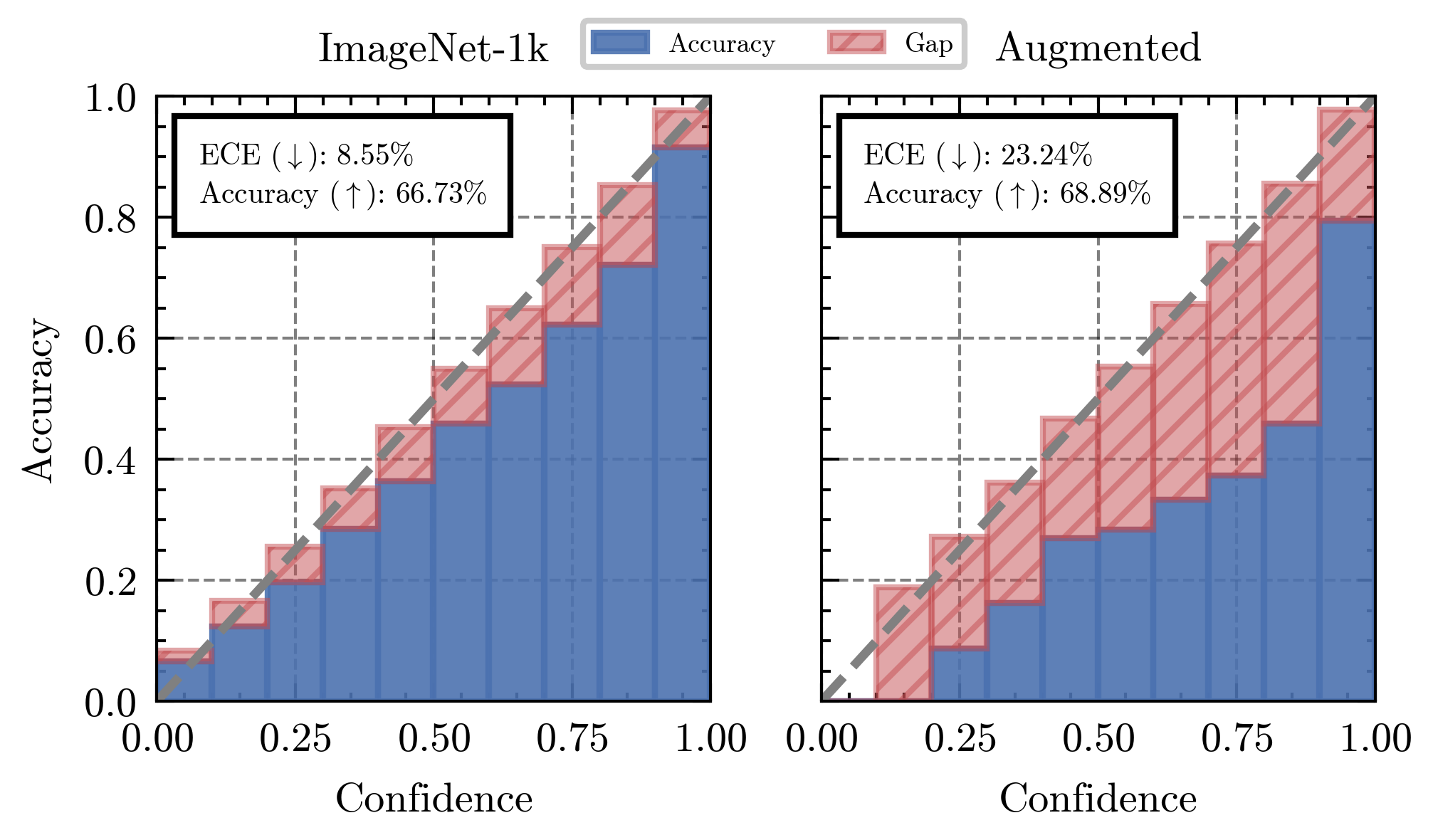}
    \caption{Reliability diagram (10 bins) for \texttt{CLIP-ViT-B-16} pretrained on LAION-2B when transferred zero-shot on ImageNet-1k. (left) Source Dataset, (right) Augmented version of the dataset.}
    \label{fig:rp_laion}
\end{figure}

%% file: tables/rpEsat.tex
\begin{wrapfigure}{o}{0.625\textwidth}
    \setlength{\abovecaptionskip}{2pt}
    \centering
    \includegraphics{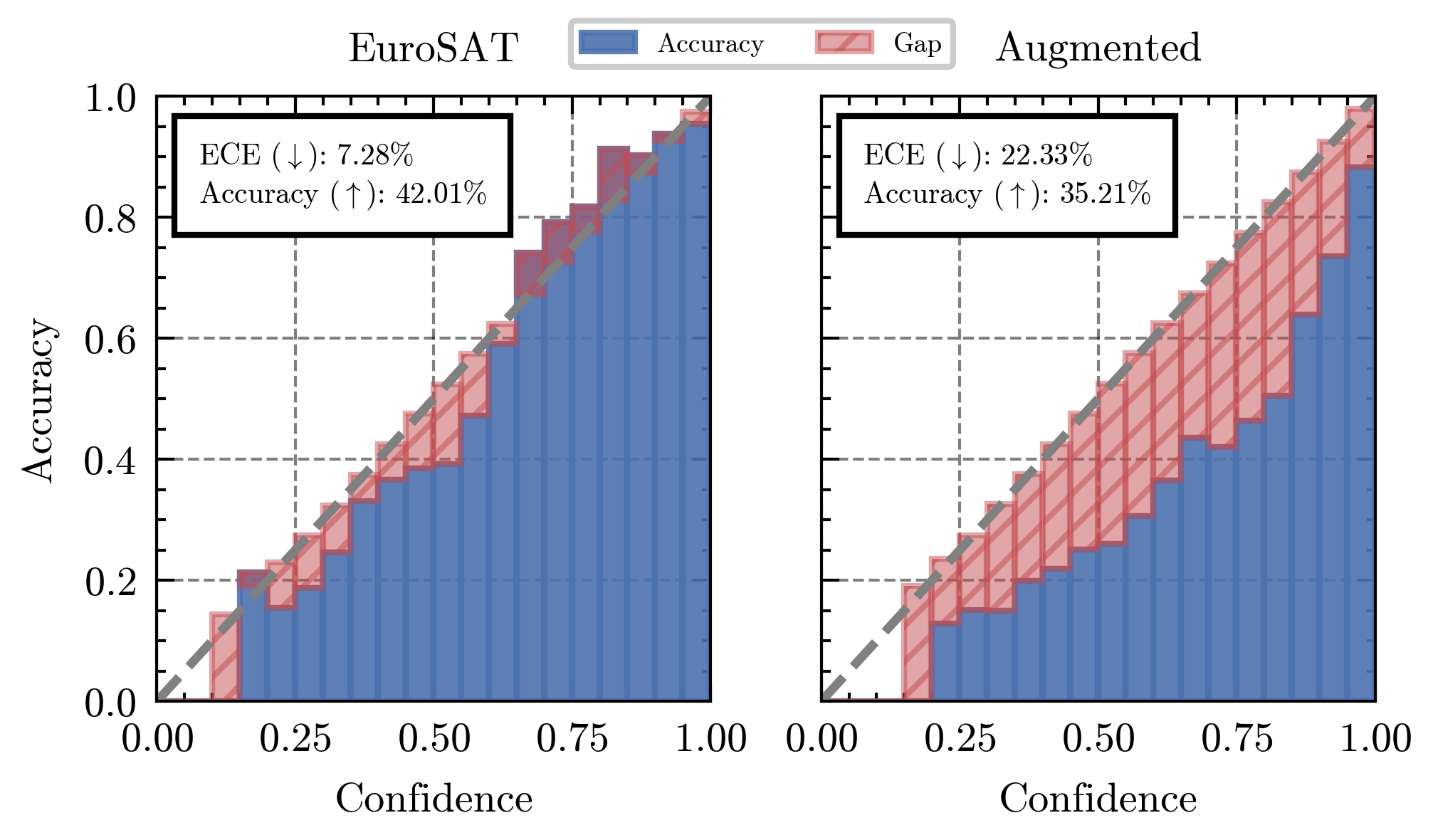}
    \caption{Reliability diagrams of \texttt{CLIP-ViT-B-16} for EuroSat and its augmented version, generated following Sec.\ref{sec:overconfidence}.}
    \label{fig:rpEsat}
\end{wrapfigure}

%% file: tables/cr/app_aug_error.tex
\begin{table}[t]
    \def\arraystretch{1.15}
    \caption{Comparison among \ding{172} CLIP's zero-shot accuracy, \ding{173} CLIP's accuracy on the augmented counterpart of the dataset, and \ding{174} \method. 
    The augmented datasets are crafted following the protocol of Section \ref{sec:overconfidence} . ``Gap'' is defined as CLIP's zero shot accuracy \emph{minus} its accuracy on the augmented dataset.
    ``Improvement'' is defined as the accuracy of \method \emph{minus} that of zero-shot CLIP. 
    Spearman's coefficient between ``Gap'' and ``Improvement'' equals $-0.95$: as the ``Gap'' decreases (\emph{i.e.}, the lower the error on augmented views) \method provides more substantial improvements. \vspace{.5em}}
    \centering
    \scriptsize
    \begin{tabularx}{\textwidth}{
    >{\raggedright\arraybackslash}p{2.5cm}
    >{\centering\arraybackslash}X
    >{\centering\arraybackslash}X
    >{\centering\arraybackslash}X
    >{\centering\arraybackslash}X
    >{\centering\arraybackslash}X
    >{\centering\arraybackslash}X
    >{\centering\arraybackslash}X
    >{\centering\arraybackslash}X
    >{\centering\arraybackslash}X
    }
    
    \toprule
    \textbf{Method} & FLWR & DTD & PETS & CARS & UCF & CAL & FOOD & SUN & AIR \\

    \midrule
     
    \ding{172} Zero-Shot & 67.44 & 44.27 & 88.25 & 65.48 & 65.13 & 93.35 & 83.65 & 62.59 & 23.67 \\

    \ding{173} Augmented & 66.19 & 44.90 & 86.17 & 65.88 & 65.59 & 92.62 & 83.25 & 62.97 & 23.52 \\
    
    \ding{174} \method (perc = $0.1$) & 67.07 & 45.80 & 86.74 &  {67.54} & 67.64 & 93.51 & 84.36 & 64.49 &  {24.40} \\

    Gap $=$ \ding{172} $-$ \ding{173} & $+$1.25 & $-$0.63 & $+$2.08 & $-$0.40 & $-$0.46 & $+$0.73 & $+$0.40 & $-$0.38 & $+$0.15 \\

    Improvement $=$ \ding{174} $-$ \ding{172} & $-$0.37 & $+$1.53 & $-$1.51 & $+$2.06 & $+$2.51 & $+$0.16 & $+$0.71 & $+$1.90 & $+$0.73 \\
    
    \bottomrule
    \end{tabularx}

    \label{tab:error_gap}
\end{table}

%% file: tables/cr/app_std.tex
\begin{table}[t!]

    \def\arraystretch{1.15}
    \centering
    \caption{Standard deviations of \method for Fine-grained classification. Each cell refers to Tab.~\ref{tab:fc}.}
    \scriptsize
    \begin{tabularx}{\textwidth}{
    >{\raggedright\arraybackslash}p{1.5cm}
    >{\centering\arraybackslash}X
    >{\centering\arraybackslash}X
    >{\centering\arraybackslash}X
    >{\centering\arraybackslash}X
    >{\centering\arraybackslash}X
    >{\centering\arraybackslash}X
    >{\centering\arraybackslash}X
    >{\centering\arraybackslash}X
    >{\centering\arraybackslash}X
    >{\centering\arraybackslash}X
    }
    
    \toprule
    \textbf{Method} & FLWR & DTD & PETS & CARS & UCF & CAL & FOOD & SUN & AIR & ESAT \\

    \midrule
    \rowcolor{lightgray}
    \multicolumn{11}{c}{\texttt{CLIP-ViT-B-16}} \\ [-2.25ex] \\

    \textsc{Zero} & $\pm$0.12 & $\pm$0.07 & $\pm$0.06 & $\pm$0.15 & $\pm$0.24 & $\pm$0.14 & $\pm$0.01 & $\pm$0.10 & $\pm$0.12 & $\pm$0.11 \\

    \textsc{Zero}\textsubscript{+Ensemble} & $\pm$0.07 & $\pm$0.26 & $\pm$0.16 & $\pm$0.04 & $\pm$0.07 & $\pm$0.19 & $\pm$0.04 & $\pm$0.18 & $\pm$0.47 & $\pm$0.08 \\

    \cmidrule(lr){1-11}
    \rowcolor{lightgray}
    \multicolumn{11}{c}{\texttt{MaPLe}} \\  [-2.25ex] \\ 
    
    \textsc{Zero} & $\pm$0.33 & $\pm$0.51 & $\pm$0.41 & $\pm$0.52 & $\pm$0.66 & $\pm$0.32 & $\pm$0.10 & $\pm$0.40 & $\pm$0.33 & $\pm$4.77 \\

    \cmidrule(lr){1-11}
    \rowcolor{lightgray}
    \multicolumn{11}{c}{\texttt{CLIP-ViT-B-16 + CLIP-ViT-L-14}} \\ [-2.25ex] \\
    
    \textsc{Zero} & $\pm$0.11 & $\pm$0.09 & $\pm$0.15 & $\pm$0.19 & $\pm$0.14 & $\pm$0.04 & $\pm$0.08 & $\pm$0.03 & $\pm$0.32 & $\pm$0.19 \\
    
    \bottomrule
    \end{tabularx}

    \label{tab:std}
\end{table}

%% file: tables/esatImages.tex
\begin{figure}[t]
    \centering
    \begin{tabular}{cccc}
        \includegraphics[width=0.22\textwidth]{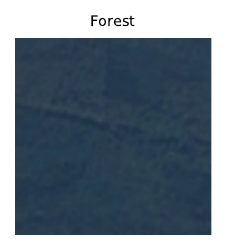} & \includegraphics[width=0.22\textwidth]{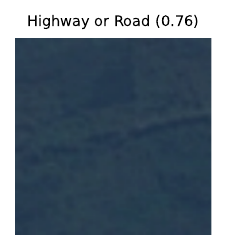} & 
        \includegraphics[width=0.22\textwidth]{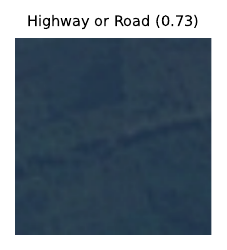} & \includegraphics[width=0.22\textwidth]{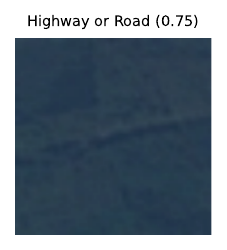} \\

        \includegraphics[width=0.235\textwidth]{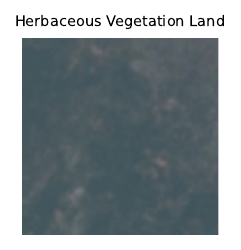} & \includegraphics[width=0.22\textwidth]{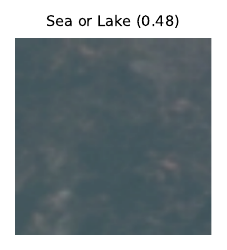} & 
        \includegraphics[width=0.22\textwidth]{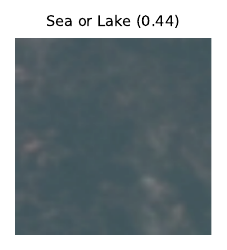} & \includegraphics[width=0.22\textwidth]{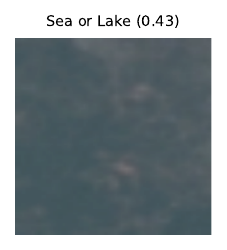} \\

        \includegraphics[width=0.22\textwidth]{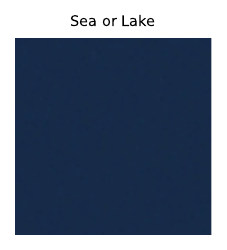} & \includegraphics[width=0.22\textwidth]{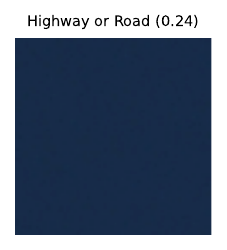} & 
        \includegraphics[width=0.22\textwidth]{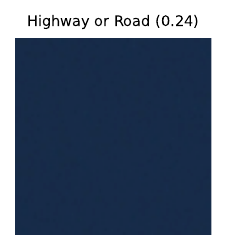} & \includegraphics[width=0.22\textwidth]{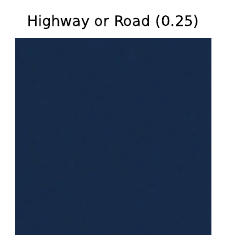} \\
        \small (a) Source Image & \small (b) Most confident & \small (c) 2nd most conf. & \small (d) 3rd most conf.
        
    \end{tabular}
    \caption{Examples of satellite imagery taken from EuroSAT\cite{helber2019eurosat}, along with augmentations leading to high confident predictions. Column (a) reports source images with their label. Columns (b-d) report views sorted by entropy (lowest to highest), paired with the prediction and the confidence of $\texttt{CLIP-ViT-B-16}$ \cite{radford2021learning}.}
    \label{fig:esat}
\end{figure}

%% file: main.bbl
\begin{thebibliography}{56}
\providecommand{\natexlab}[1]{#1}
\providecommand{\url}[1]{\texttt{#1}}
\expandafter\ifx\csname urlstyle\endcsname\relax
  \providecommand{\doi}[1]{doi: #1}\else
  \providecommand{\doi}{doi: \begingroup \urlstyle{rm}\Url}\fi

\bibitem[Bossard et~al.(2014)Bossard, Guillaumin, and Van~Gool]{bossard2014food}
Lukas Bossard, Matthieu Guillaumin, and Luc Van~Gool.
\newblock Food-101--mining discriminative components with random forests.
\newblock In \emph{European Conference on Computer Vision (ECCV)}, 2014.

\bibitem[Cherti et~al.(2023)Cherti, Beaumont, Wightman, Wortsman, Ilharco, Gordon, Schuhmann, Schmidt, and Jitsev]{cherti2023reproducible}
Mehdi Cherti, Romain Beaumont, Ross Wightman, Mitchell Wortsman, Gabriel Ilharco, Cade Gordon, Christoph Schuhmann, Ludwig Schmidt, and Jenia Jitsev.
\newblock Reproducible scaling laws for contrastive language-image learning.
\newblock In \emph{IEEE/CVF Conference on Computer Vision and Pattern Recognition (CVPR)}, 2023.

\bibitem[Cimpoi et~al.(2014)Cimpoi, Maji, Kokkinos, Mohamed, and Vedaldi]{cimpoi2014describing}
Mircea Cimpoi, Subhransu Maji, Iasonas Kokkinos, Sammy Mohamed, and Andrea Vedaldi.
\newblock Describing textures in the wild.
\newblock In \emph{IEEE/CVF Conference on Computer Vision and Pattern Recognition (CVPR)}, 2014.

\bibitem[Deng et~al.(2009)Deng, Dong, Socher, Li, Li, and Fei-Fei]{deng2009imagenet}
Jia Deng, Wei Dong, Richard Socher, Li-Jia Li, Kai Li, and Li~Fei-Fei.
\newblock Imagenet: A large-scale hierarchical image database.
\newblock In \emph{IEEE/CVF Conference on Computer Vision and Pattern Recognition (CVPR)}, 2009.

\bibitem[Dosovitskiy et~al.(2020)Dosovitskiy, Beyer, Kolesnikov, Weissenborn, Zhai, Unterthiner, Dehghani, Minderer, Heigold, Gelly, et~al.]{dosovitskiy2020image}
Alexey Dosovitskiy, Lucas Beyer, Alexander Kolesnikov, Dirk Weissenborn, Xiaohua Zhai, Thomas Unterthiner, Mostafa Dehghani, Matthias Minderer, Georg Heigold, Sylvain Gelly, et~al.
\newblock An image is worth 16x16 words: Transformers for image recognition at scale.
\newblock In \emph{International Conference on Learning Representations (ICLR)}, 2020.

\bibitem[Fei-Fei et~al.(2004)Fei-Fei, Fergus, and Perona]{fei2004learning}
Li~Fei-Fei, Rob Fergus, and Pietro Perona.
\newblock Learning generative visual models from few training examples: An incremental bayesian approach tested on 101 object categories.
\newblock In \emph{IEEE/CVF Conference on Computer Vision and Pattern Recognition Workshops (CVPR-W)}. IEEE, 2004.

\bibitem[Gandelsman et~al.(2022)Gandelsman, Sun, Chen, and Efros]{gandelsman2022test}
Yossi Gandelsman, Yu~Sun, Xinlei Chen, and Alexei Efros.
\newblock Test-time training with masked autoencoders.
\newblock \emph{Advances in Neural Information Processing Systems (NeurIPS)}, 2022.

\bibitem[Guo et~al.(2017)Guo, Pleiss, Sun, and Weinberger]{guo2017calibration}
Chuan Guo, Geoff Pleiss, Yu~Sun, and Kilian~Q Weinberger.
\newblock On calibration of modern neural networks.
\newblock In \emph{International Conference on Machine Learning (ICML)}, 2017.

\bibitem[Hardt and Sun(2024)]{hardt2024testtime}
Moritz Hardt and Yu~Sun.
\newblock Test-time training on nearest neighbors for large language models.
\newblock In \emph{International Conference on Learning Representations (ICLR)}, 2024.

\bibitem[He et~al.(2016)He, Zhang, Ren, and Sun]{he2016deep}
Kaiming He, Xiangyu Zhang, Shaoqing Ren, and Jian Sun.
\newblock Deep residual learning for image recognition.
\newblock In \emph{IEEE/CVF Conference on Computer Vision and Pattern Recognition (CVPR)}, 2016.

\bibitem[Helber et~al.(2019)Helber, Bischke, Dengel, and Borth]{helber2019eurosat}
Patrick Helber, Benjamin Bischke, Andreas Dengel, and Damian Borth.
\newblock Eurosat: A novel dataset and deep learning benchmark for land use and land cover classification.
\newblock \emph{IEEE Journal of Selected Topics in Applied Earth Observations and Remote Sensing}, 12\penalty0 (7), 2019.

\bibitem[Hendrycks et~al.(2021{\natexlab{a}})Hendrycks, Basart, Mu, Kadavath, Wang, Dorundo, Desai, Zhu, Parajuli, Guo, et~al.]{hendrycks2021many}
Dan Hendrycks, Steven Basart, Norman Mu, Saurav Kadavath, Frank Wang, Evan Dorundo, Rahul Desai, Tyler Zhu, Samyak Parajuli, Mike Guo, et~al.
\newblock The many faces of robustness: A critical analysis of out-of-distribution generalization.
\newblock In \emph{IEEE/CVF International Conference on Computer Vision (ICCV)}, 2021{\natexlab{a}}.

\bibitem[Hendrycks et~al.(2021{\natexlab{b}})Hendrycks, Zhao, Basart, Steinhardt, and Song]{hendrycks2021natural}
Dan Hendrycks, Kevin Zhao, Steven Basart, Jacob Steinhardt, and Dawn Song.
\newblock Natural adversarial examples.
\newblock In \emph{IEEE/CVF Conference on Computer Vision and Pattern Recognition (CVPR)}, 2021{\natexlab{b}}.

\bibitem[Jia et~al.(2021)Jia, Yang, Xia, Chen, Parekh, Pham, Le, Sung, Li, and Duerig]{jia2021scaling}
Chao Jia, Yinfei Yang, Ye~Xia, Yi-Ting Chen, Zarana Parekh, Hieu Pham, Quoc Le, Yun-Hsuan Sung, Zhen Li, and Tom Duerig.
\newblock Scaling up visual and vision-language representation learning with noisy text supervision.
\newblock In \emph{International Conference on Machine Learning (ICML)}, 2021.

\bibitem[Khattak et~al.(2023)Khattak, Rasheed, Maaz, Khan, and Khan]{khattak2023maple}
Muhammad~Uzair Khattak, Hanoona Rasheed, Muhammad Maaz, Salman Khan, and Fahad~Shahbaz Khan.
\newblock Maple: Multi-modal prompt learning.
\newblock In \emph{IEEE/CVF Conference on Computer Vision and Pattern Recognition (CVPR)}, 2023.

\bibitem[Kimura(2021)]{kimura2021understanding}
Masanari Kimura.
\newblock Understanding test-time augmentation.
\newblock In \emph{International Conference on Neural Information Processing (ICONIP)}. Springer, 2021.

\bibitem[Krause et~al.(2013)Krause, Stark, Deng, and Fei-Fei]{krause20133d}
Jonathan Krause, Michael Stark, Jia Deng, and Li~Fei-Fei.
\newblock 3d object representations for fine-grained categorization.
\newblock In \emph{IEEE/CVF International Conference on Computer Vision Workshops (ICCV-W)}, 2013.

\bibitem[Kuncheva(2014)]{kuncheva2014combining}
Ludmila~I Kuncheva.
\newblock \emph{Combining pattern classifiers: methods and algorithms}.
\newblock 2014.

\bibitem[Lee et~al.(2024)Lee, Jung, Lee, Park, Shin, Hwang, and Yoon]{lee2024entropy}
Jonghyun Lee, Dahuin Jung, Saehyung Lee, Junsung Park, Juhyeon Shin, Uiwon Hwang, and Sungroh Yoon.
\newblock Entropy is not enough for test-time adaptation: From the perspective of disentangled factors.
\newblock In \emph{International Conference on Learning Representations (ICLR)}, 2024.

\bibitem[Li and Liang(2021)]{li2021prefix}
Xiang~Lisa Li and Percy Liang.
\newblock Prefix-tuning: Optimizing continuous prompts for generation.
\newblock In \emph{Proceedings of the 59th Annual Meeting of the Association for Computational Linguistics and the 11th International Joint Conference on Natural Language Processing (Volume 1: Long Papers)}, 2021.

\bibitem[Liu et~al.(2024)Liu, Sun, Peng, and Zhou]{liu2024dart}
Zichen Liu, Hongbo Sun, Yuxin Peng, and Jiahuan Zhou.
\newblock Dart: Dual-modal adaptive online prompting and knowledge retention for test-time adaptation.
\newblock In \emph{AAAI Conference on Artificial Intelligence (AAAI)}, 2024.

\bibitem[Ma et~al.(2024)Ma, Zhang, Guo, and Xu]{ma2024swapprompt}
Xiaosong Ma, Jie Zhang, Song Guo, and Wenchao Xu.
\newblock Swapprompt: Test-time prompt adaptation for vision-language models.
\newblock \emph{Advances in Neural Information Processing Systems (NeurIPS)}, 2024.

\bibitem[Maji et~al.(2013)Maji, Rahtu, Kannala, Blaschko, and Vedaldi]{maji2013fine}
Subhransu Maji, Esa Rahtu, Juho Kannala, Matthew Blaschko, and Andrea Vedaldi.
\newblock Fine-grained visual classification of aircraft.
\newblock \emph{arXiv preprint arXiv:1306.5151}, 2013.

\bibitem[Mayilvahanan et~al.(2023)Mayilvahanan, Wiedemer, Rusak, Bethge, and Brendel]{mayilvahanan2023does}
Prasanna Mayilvahanan, Thadd{\"a}us Wiedemer, Evgenia Rusak, Matthias Bethge, and Wieland Brendel.
\newblock Does clip’s generalization performance mainly stem from high train-test similarity?
\newblock In \emph{International Conference on Learning Representations (ICLR)}, 2023.

\bibitem[Nilsback and Zisserman(2008)]{nilsback2008automated}
Maria-Elena Nilsback and Andrew Zisserman.
\newblock Automated flower classification over a large number of classes.
\newblock In \emph{Indian conference on computer vision, graphics \& image processing}. IEEE, 2008.

\bibitem[Niu et~al.(2022)Niu, Wu, Zhang, Chen, Zheng, Zhao, and Tan]{niu2022efficient}
Shuaicheng Niu, Jiaxiang Wu, Yifan Zhang, Yaofo Chen, Shijian Zheng, Peilin Zhao, and Mingkui Tan.
\newblock Efficient test-time model adaptation without forgetting.
\newblock In \emph{International Conference on Machine Learning (ICML)}, 2022.

\bibitem[Niu et~al.(2023)Niu, Wu, Zhang, Wen, Chen, Zhao, and Tan]{niu2023towards}
Shuaicheng Niu, Jiaxiang Wu, Yifan Zhang, Zhiquan Wen, Yaofo Chen, Peilin Zhao, and Mingkui Tan.
\newblock Towards stable test-time adaptation in dynamic wild world.
\newblock In \emph{International Conference on Learning Representations (ICLR)}, 2023.

\bibitem[OpenAI()]{openai2024clip}
OpenAI.
\newblock Clip.
\newblock URL \url{https://github.com/openai/CLIP}.

\bibitem[Parkhi et~al.(2012)Parkhi, Vedaldi, Zisserman, and Jawahar]{parkhi2012cats}
Omkar~M Parkhi, Andrea Vedaldi, Andrew Zisserman, and CV~Jawahar.
\newblock Cats and dogs.
\newblock In \emph{IEEE/CVF Conference on Computer Vision and Pattern Recognition (CVPR)}, 2012.

\bibitem[Paszke et~al.(2019)Paszke, Gross, Massa, Lerer, Bradbury, Chanan, Killeen, Lin, Gimelshein, Antiga, et~al.]{paszke2019pytorch}
Adam Paszke, Sam Gross, Francisco Massa, Adam Lerer, James Bradbury, Gregory Chanan, Trevor Killeen, Zeming Lin, Natalia Gimelshein, Luca Antiga, et~al.
\newblock Pytorch: An imperative style, high-performance deep learning library.
\newblock \emph{Advances in Neural Information Processing Systems (NeurIPS)}, 32, 2019.

\bibitem[Radford et~al.(2021)Radford, Kim, Hallacy, Ramesh, Goh, Agarwal, Sastry, Askell, Mishkin, Clark, et~al.]{radford2021learning}
Alec Radford, Jong~Wook Kim, Chris Hallacy, Aditya Ramesh, Gabriel Goh, Sandhini Agarwal, Girish Sastry, Amanda Askell, Pamela Mishkin, Jack Clark, et~al.
\newblock Learning transferable visual models from natural language supervision.
\newblock In \emph{International Conference on Machine Learning (ICML)}, 2021.

\bibitem[Recht et~al.(2019)Recht, Roelofs, Schmidt, and Shankar]{recht2019imagenet}
Benjamin Recht, Rebecca Roelofs, Ludwig Schmidt, and Vaishaal Shankar.
\newblock Do imagenet classifiers generalize to imagenet?
\newblock In \emph{International Conference on Machine Learning (ICML)}, 2019.

\bibitem[Samadh et~al.(2023)Samadh, Gani, Hussein, Khattak, Naseer, Khan, and Khan]{samadh2023align}
Jameel Hassan~Abdul Samadh, Hanan Gani, Noor~Hazim Hussein, Muhammad~Uzair Khattak, Muzammal Naseer, Fahad Khan, and Salman Khan.
\newblock Align your prompts: Test-time prompting with distribution alignment for zero-shot generalization.
\newblock In \emph{Advances in Neural Information Processing Systems (NeurIPS)}, 2023.

\bibitem[Schuhmann et~al.(2022)Schuhmann, Beaumont, Vencu, Gordon, Wightman, Cherti, Coombes, Katta, Mullis, Wortsman, et~al.]{schuhmann2022laion}
Christoph Schuhmann, Romain Beaumont, Richard Vencu, Cade Gordon, Ross Wightman, Mehdi Cherti, Theo Coombes, Aarush Katta, Clayton Mullis, Mitchell Wortsman, et~al.
\newblock Laion-5b: An open large-scale dataset for training next generation image-text models.
\newblock \emph{Advances in Neural Information Processing Systems (NeurIPS)}, 2022.

\bibitem[Shanmugam et~al.(2021)Shanmugam, Blalock, Balakrishnan, and Guttag]{shanmugam2021better}
Divya Shanmugam, Davis Blalock, Guha Balakrishnan, and John Guttag.
\newblock Better aggregation in test-time augmentation.
\newblock In \emph{IEEE/CVF International Conference on Computer Vision (ICCV)}, 2021.

\bibitem[Shapley and Grofman(1984)]{shapley1984optimizing}
Lloyd Shapley and Bernard Grofman.
\newblock Optimizing group judgmental accuracy in the presence of interdependencies.
\newblock \emph{Public Choice}, 1984.

\bibitem[Sharma et~al.(2018)Sharma, Ding, Goodman, and Soricut]{sharma2018conceptual}
Piyush Sharma, Nan Ding, Sebastian Goodman, and Radu Soricut.
\newblock Conceptual captions: A cleaned, hypernymed, image alt-text dataset for automatic image captioning.
\newblock In \emph{Proceedings of the 56th Annual Meeting of the Association for Computational Linguistics (Volume 1: Long Papers)}, 2018.

\bibitem[Shu et~al.(2022)Shu, Nie, Huang, Yu, Goldstein, Anandkumar, and Xiao]{shu2022test}
Manli Shu, Weili Nie, De-An Huang, Zhiding Yu, Tom Goldstein, Anima Anandkumar, and Chaowei Xiao.
\newblock Test-time prompt tuning for zero-shot generalization in vision-language models.
\newblock \emph{Advances in Neural Information Processing Systems (NeurIPS)}, 2022.

\bibitem[Singh et~al.(2022)Singh, Hu, Goswami, Couairon, Galuba, Rohrbach, and Kiela]{singh2022flava}
Amanpreet Singh, Ronghang Hu, Vedanuj Goswami, Guillaume Couairon, Wojciech Galuba, Marcus Rohrbach, and Douwe Kiela.
\newblock Flava: A foundational language and vision alignment model.
\newblock In \emph{IEEE/CVF Conference on Computer Vision and Pattern Recognition (CVPR)}, 2022.

\bibitem[Son and Kang(2023)]{son2023efficient}
Jongwook Son and Seokho Kang.
\newblock Efficient improvement of classification accuracy via selective test-time augmentation.
\newblock \emph{Information Sciences}, 2023.

\bibitem[Soomro et~al.(2012)Soomro, Zamir, and Shah]{soomro2012ucf101}
Khurram Soomro, Amir~Roshan Zamir, and Mubarak Shah.
\newblock Ucf101: A dataset of 101 human actions classes from videos in the wild.
\newblock \emph{arXiv preprint arXiv:1212.0402}, 2012.

\bibitem[Sui et~al.(2024)Sui, Wang, and Yeung-Levy]{sui2024just}
Elaine Sui, Xiaohan Wang, and Serena Yeung-Levy.
\newblock Just shift it: Test-time prototype shifting for zero-shot generalization with vision-language models.
\newblock \emph{arXiv preprint arXiv:2403.12952}, 2024.

\bibitem[Sun et~al.(2020)Sun, Wang, Liu, Miller, Efros, and Hardt]{sun2020test}
Yu~Sun, Xiaolong Wang, Zhuang Liu, John Miller, Alexei Efros, and Moritz Hardt.
\newblock Test-time training with self-supervision for generalization under distribution shifts.
\newblock In \emph{International Conference on Machine Learning (ICML)}, 2020.

\bibitem[Tomar et~al.(2023)Tomar, Vray, Bozorgtabar, and Thiran]{tomar2023tesla}
Devavrat Tomar, Guillaume Vray, Behzad Bozorgtabar, and Jean-Philippe Thiran.
\newblock Tesla: Test-time self-learning with automatic adversarial augmentation.
\newblock In \emph{IEEE/CVF Conference on Computer Vision and Pattern Recognition (CVPR)}, 2023.

\bibitem[Wang et~al.(2021)Wang, Shelhamer, Liu, Olshausen, and Darrell]{wang2020tent}
Dequan Wang, Evan Shelhamer, Shaoteng Liu, Bruno Olshausen, and Trevor Darrell.
\newblock Tent: Fully test-time adaptation by entropy minimization.
\newblock In \emph{International Conference on Learning Representations (ICLR)}, 2021.

\bibitem[Wang et~al.(2019)Wang, Ge, Lipton, and Xing]{wang2019learning}
Haohan Wang, Songwei Ge, Zachary Lipton, and Eric~P Xing.
\newblock Learning robust global representations by penalizing local predictive power.
\newblock \emph{Advances in Neural Information Processing Systems (NeurIPS)}, 2019.

\bibitem[Wang et~al.(2022)Wang, Fink, Van~Gool, and Dai]{wang2022continual}
Qin Wang, Olga Fink, Luc Van~Gool, and Dengxin Dai.
\newblock Continual test-time domain adaptation.
\newblock In \emph{IEEE/CVF Conference on Computer Vision and Pattern Recognition (CVPR)}, 2022.

\bibitem[Xiao et~al.(2010)Xiao, Hays, Ehinger, Oliva, and Torralba]{xiao2010sun}
Jianxiong Xiao, James Hays, Krista~A Ehinger, Aude Oliva, and Antonio Torralba.
\newblock Sun database: Large-scale scene recognition from abbey to zoo.
\newblock In \emph{IEEE/CVF Conference on Computer Vision and Pattern Recognition (CVPR)}, 2010.

\bibitem[Yuan et~al.(2023)Yuan, Xie, and Li]{yuan2023robust}
Longhui Yuan, Binhui Xie, and Shuang Li.
\newblock Robust test-time adaptation in dynamic scenarios.
\newblock In \emph{IEEE/CVF Conference on Computer Vision and Pattern Recognition (CVPR)}, 2023.

\bibitem[Zancato et~al.(2023)Zancato, Achille, Liu, Trager, Perera, and Soatto]{zancato2023train}
Luca Zancato, Alessandro Achille, Tian~Yu Liu, Matthew Trager, Pramuditha Perera, and Stefano Soatto.
\newblock Train/test-time adaptation with retrieval.
\newblock In \emph{IEEE/CVF Conference on Computer Vision and Pattern Recognition (CVPR)}, 2023.

\bibitem[Zanella and Ayed(2024)]{zanella2024test}
Maxime Zanella and Ismail~Ben Ayed.
\newblock On the test-time zero-shot generalization of vision-language models: Do we really need prompt learning?
\newblock \emph{arXiv preprint arXiv:2405.02266}, 2024.

\bibitem[Zhai et~al.(2023)Zhai, Mustafa, Kolesnikov, and Beyer]{zhai2023sigmoid}
Xiaohua Zhai, Basil Mustafa, Alexander Kolesnikov, and Lucas Beyer.
\newblock Sigmoid loss for language image pre-training.
\newblock In \emph{IEEE/CVF International Conference on Computer Vision (ICCV)}, 2023.

\bibitem[Zhang et~al.(2022)Zhang, Levine, and Finn]{zhang2022memo}
Marvin Zhang, Sergey Levine, and Chelsea Finn.
\newblock Memo: Test time robustness via adaptation and augmentation.
\newblock \emph{Advances in Neural Information Processing Systems (NeurIPS)}, 2022.

\bibitem[Zhao et~al.(2024)Zhao, Wang, Zhu, and Yang]{zhao2024testtime}
Shuai Zhao, Xiaohan Wang, Linchao Zhu, and Yi~Yang.
\newblock Test-time adaptation with {CLIP} reward for zero-shot generalization in vision-language models.
\newblock In \emph{International Conference on Learning Representations (ICLR)}, 2024.

\bibitem[Zhou et~al.(2022{\natexlab{a}})Zhou, Yang, Loy, and Liu]{zhou2022conditional}
Kaiyang Zhou, Jingkang Yang, Chen~Change Loy, and Ziwei Liu.
\newblock Conditional prompt learning for vision-language models.
\newblock In \emph{IEEE/CVF Conference on Computer Vision and Pattern Recognition (CVPR)}, 2022{\natexlab{a}}.

\bibitem[Zhou et~al.(2022{\natexlab{b}})Zhou, Yang, Loy, and Liu]{zhou2022learning}
Kaiyang Zhou, Jingkang Yang, Chen~Change Loy, and Ziwei Liu.
\newblock Learning to prompt for vision-language models.
\newblock \emph{International Journal of Computer Vision (IJCV)}, 2022{\natexlab{b}}.

\end{thebibliography}
